\documentclass[runningheads]{llncs}

\usepackage{eccv}

\usepackage{eccvabbrv}
\usepackage{appendix}

\usepackage{graphicx}
\usepackage{booktabs}

\usepackage[accsupp]{axessibility}  % Improves PDF readability for those with disabilities.

\usepackage{orcidlink}

\usepackage{hyperref}

\usepackage{multirow}
\newcommand{\myparagraph}[1]{
\noindent \textbf{#1} ---
}

\newcommand{\ours}{CAR-MIL\xspace}

\usepackage[table]{xcolor}
\definecolor{Yellow}{RGB}{220,204,118}
\newcommand{\our}{\cellcolor{Yellow!15}}

\newcommand{\pos}[1]{\textcolor{black}{#1}}
\newcommand{\ic}[1]{\textcolor{black}{#1}}

\begin{document}

% ---------------------------------------------------------------
% TODO REVIEW: Replace with your title
\title{CAR-MIL: Counterfactual Attention Regularization for Multiple Instance Learning} 
 
% TODO REVIEW: If the paper title is too long for the running head, you can set
% an abbreviated paper title here. If not, comment out.
\titlerunning{CAR-MIL}

% TODO FINAL: Replace with your author list. 
% Include the authors' OCRID for the camera-ready version, if at all possible.
\author{Imane Chraki\inst{1,2} \and
Pierre Marza\inst{1,2} \and
Stergios Christodoulidis\inst{1,2} \and
Maria Vakalopoulou\inst{1,2}}
% TODO FINAL: Replace with an abbreviated list of authors.
\authorrunning{I.Chraki et al.}
% First names are abbreviated in the running head.
% If there are more than two authors, 'et al.' is used.

% TODO FINAL: Replace with your institution list.
\institute{Université Paris-Saclay, CentraleSupélec, Gustave Roussy, INSERM, IHU PRISM, Cancer Data Science Unit, France \and
Université Paris-Saclay, CentraleSupélec, MICS Laboratory, France
% \email{lncs@springer.com} 
}

\maketitle

\begin{abstract}
\label{sec:abstract}

Multiple Instance Learning (MIL) is widely used for weakly supervised learning, particularly in digital pathology, where fine-grained annotations are costly. Most MIL methods aggregate instance features via attention mechanisms. However, attention weights do not always faithfully reflect instance importance and may focus on spuriously correlated regions. In this work, we propose \ours, a framework that explicitly guides attention learning through a counterfactual attention regularization objective inspired by counterfactual explanations. Built on a standard attention-based MIL architecture, our approach introduces a lightweight counterfactual attention branch trained to produce an alternative prediction while remaining close to the factual attention distribution. This encourages prediction changes to arise from minimal, structured redistributions of attention, leading to more informative evidence allocation. The resulting factual and counterfactual attention maps capture complementary evidence: the former highlights regions supporting the prediction, while the latter reveals regions whose reweighting would challenge it. We evaluate our method on synthetic MIL benchmarks with instance-level ground truth enabling controlled analysis of attention behavior and on five digital pathology datasets across four tasks. \ours \pos{maintains competitive classification performance, with the largest gains observed on more challenging tasks, while improving attention reliability, demonstrating the benefits of integrating counterfactual explainability reasoning into attention learning.} 
Code is available at: \url{https://github.com/ImaneCR/CAR-MIL/}.
% \ic{consistently improves classification performance} and attention reliability, demonstrating the benefit of integrating counterfactual explainability into attention learning.
% \keywords{First keyword \and Second keyword \and Third keyword}
\end{abstract}

\section{Introduction}
\label{sec:intro}

\begin{figure}[t]
    \centering
    \includegraphics[width=\linewidth]{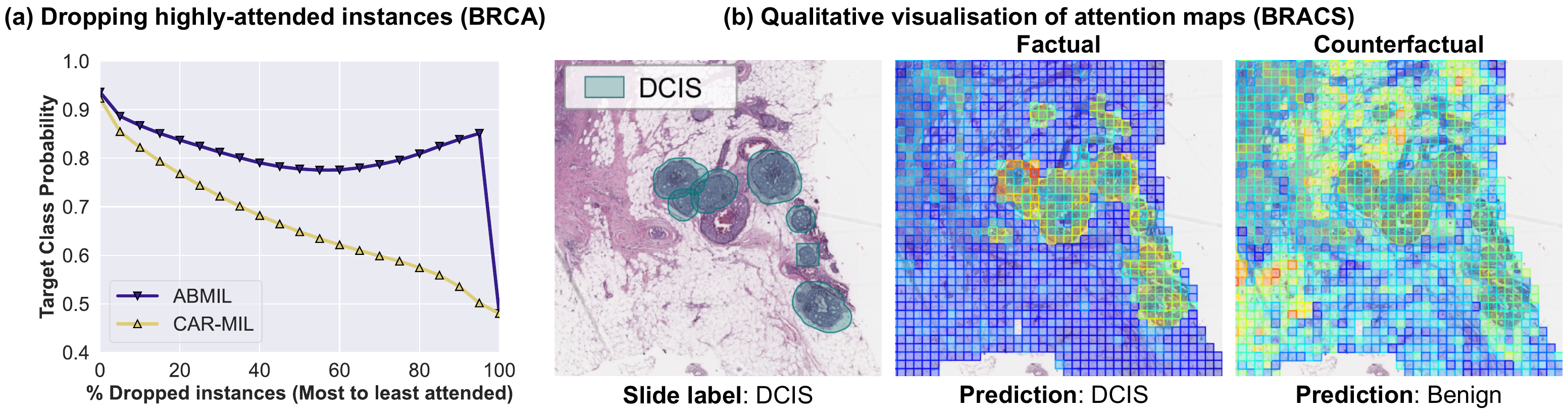}
    \caption{\textbf{\ours} is a multiple instance learning method to model complex interactions between relevant regions using counterfactual explanation.
    \textbf{(a)} Dropping highly-attended instances for CAR-MIL leads to a higher performance drop than ABMIL on the TCGA-BRCA dataset.
    %where vanilla ABMIL showcases a failure mode. 
    \textbf{(b)} Qualitative example from the BRACS dataset: the factual attention correctly focuses on the \textit{Ductal Carcinoma In Situ} (\textit{DCIS}) areas and predicts the correct slide-level label, whereas the counterfactual attention reallocates evidence and predicts the \textit{Benign} class.}    \label{fig:teaser}
\end{figure}

Multiple Instance Learning (MIL) addresses learning from weakly-annotated data, where samples are organized into bags of instances, and labels are assigned only at the bag level  \cite{MIL,mil_maron1997framework}. 

Attention-based MIL methods \cite{ilse2018attentionbaseddeepmultipleinstance,lu2021ai,wagner2023transformer} address this challenge by learning a weighted aggregation of instance features into a bag representation, and achieve high predictive performance.
Despite recent progress, two major limitations remain. First, as shown in recent studies~\cite{hense2025xmilinsightfulexplanationsmultiple,zhang2022dtfdmildoubletierfeaturedistillation,early2024inherentlyinterpretabletimeseries,javed2022additivemilintrinsicallyinterpretable}, MIL attention may highlight irrelevant or spuriously-correlated regions, both degrading performance and interpretability. Such an issue can arise from training instability, with the attention collapsing and highlighting only a small set of regions \cite{zhang2024attentionchallengingmultipleinstancelearning,mhim-mil}. Secondly, current methods only emphasize evidence supporting the predicted class while providing little insight into evidence that contradicts the decision, limiting both their reliability and their explanatory capacity. This is for example problematic in medical applications, where attention maps are often used to justify automated decisions.
These observations suggest that attention should not be treated merely as a by-product of optimization but its training should be guided to reflect how evidence supporting or contradicting a prediction is allocated across instances.

A natural framework for reasoning about such evidence is provided by counterfactual explanations (CE) \cite{wachter2018counterfactualexplanationsopeningblack}, which characterize predictions through minimal changes to the input of a model that would alter its output. This counterfactual perspective distinguishes evidence supporting a decision from evidence that challenges it. Importantly, existing CE methods provide post-hoc interpretability, as they are applied to a trained model to better understand its decision-making process. However, as showcased earlier, in the case of MIL, evidence reasoning should be performed at training time. We thus propose two main adaptations to this framework: (i) integrating CE at training time to guide the learning of attention, (ii) performing perturbations in attention instead of input space as it is more adapted to MIL.
To the best of our knowledge, such an adaptation of CE to MIL training has not been studied in previous work. Recent methods begin to incorporate counterfactual interventions into the training of attention \cite{rao2021counterfactualattentionlearningfinegrained, chraki2026counterfactual} but this is different from what we propose. Indeed, the latter rely on random attention perturbations while we believe learning such perturbations is beneficial. 

In this work, we introduce \ours (Figure~\ref{fig:teaser}), a method trained to provide a factual and counterfactual attentions following CE principles. The counterfactual attention is constrained to remain close to the factual attention while producing a different prediction, encouraging changes to arise from selective redistributions of attention across instances. We propose a double-branch design yielding complementary attention maps: the factual attention highlights regions supporting the prediction, while the counterfactual attention reveals regions refuting it. Our framework introduces a new learning paradigm where the model learns how reallocating attention across instances affects the predicted outcome and therefore providing a contrastive training signal that guides attention learning dynamics. The counterfactual branch is implemented as a lightweight attention module sharing the encoder and classifier with the factual branch, therefore introducing minimal computational overhead and requiring no additional supervision. 
We evaluate \ours on synthetic MIL benchmarks designed to assess interaction modelling \cite{hense2025xmilinsightfulexplanationsmultiple}, as well as on five digital pathology datasets across four tasks. 
\pos{Across all settings, our method consistently maintains  a competitive classification performance, with the largest performance gains observed on the more challenging tasks, while producing more reliable attention distributions, as confirmed by both quantitative metrics and qualitative analysis.}

Our contributions can be summarized as follows:
(i) We introduce for the first time counterfactual attention regularization for MIL, enforcing prediction divergence under minimal attention perturbations.
(ii) We learn complementary factual and counterfactual attention distributions that reveal supporting and refuting evidence within attention maps.
(iii) \pos{We demonstrate that explicitly guiding attention during training maintains or improves both predictive accuracy and interpretability in attention-based MIL models.}

\section{Related Work}
\label{sec:related_work}

\myparagraph{Attention-based embedding-level MIL} Such MIL models aggregate instance embeddings into a global bag representation through an attention mechanism.
The standard attention MIL (ABMIL) method \cite{ilse2018attentionbaseddeepmultipleinstance} infers weights from instance features to aggregate them accordingly.

Numerous variants have then been proposed to improve the attention module~\cite{shao2021transmiltransformerbasedcorrelated, clam,li2021dualstreammultipleinstancelearning, mhim-mil, qu2022bidirectionalweaklysupervisedknowledge,zhang2022dtfdmildoubletierfeaturedistillation, zhang2024attentionchallengingmultipleinstancelearning}.
%To improve the attention module of the MIL framework, numerous variants have followed 
 Many of these MIL variants aim to improve the attention learning process, for example, by incorporating self-attention mechanisms \cite{xiong2021nystromformernystrombasedalgorithmapproximating}, enforcing consistency between attention weights and instance feature representations \cite{clam}, constraining attention through negative bag supervision in the case of binary classification \cite{qu2022bidirectionalweaklysupervisedknowledge}, or mitigating problems caused by hard instances by introducing masking modules \cite{mhim-mil}. However, modifying only the attention mechanism, without evaluating the impact of the applied techniques on the model’s predictions, might not guarantee achieving the desired predictive outcome \cite{rao2021counterfactualattentionlearningfinegrained}.

\myparagraph{Interpretability of MIL Models} Attention-based MIL models rely on attention scores to provide a localization map of regions of interest. However, raw attention maps inherent to these models often do not guarantee reliability \cite{hense2025xmilinsightfulexplanationsmultiple, javed2022additivemilintrinsicallyinterpretable, zhang2022dtfdmildoubletierfeaturedistillation}. On the other hand, fully additive models such as \cite{javed2022additivemilintrinsicallyinterpretable} seek to disentangle and sum individual contributions explicitly. Post-hoc explainability strategies have also been proposed \cite{ adebayo2018sanity,AI_reliability,histo_interp_1, histo_interp_2, bach2015pixel,baehrens2010explain, shrikumar2017learning, montavon2019layer, hense2025xmilinsightfulexplanationsmultiple}, including perturbation-based methods that modify instance subsets within bags and observe prediction changes \cite{early2024inherentlyinterpretabletimeseries,hense2025xmilinsightfulexplanationsmultiple}.  Despite progress, many of these methods are constrained by computational cost, limited scalability to large bags \cite{molnar2020general, van2022tractability, javed2022additivemilintrinsicallyinterpretable} or limited explanatory capacity: they typically highlight instances that support the predicted class, but do not reveal evidence that contradicts or challenges the decision \cite{hense2025xmilinsightfulexplanationsmultiple}. 

\myparagraph{Counterfactual Reasoning for Explanation and Attention Learning} We can consider two main frameworks with different objectives: (i) Counterfactual Explainability (CE) tries to understand the inner mechanisms of a trained model, while (ii) Counterfactual Attention Learning (CAL) aims at improving the quality of attention at training time.
CE was introduced by \cite{wachter2018counterfactualexplanationsopeningblack} as a post-hoc interpretability framework that explains a model’s prediction by identifying minimal changes to the input that would alter its decision. The core idea is to characterize predictions through contrast: what needs to change for the outcome to differ. CE methods therefore operate after training and focus on generating alternative inputs that provide insight into decision boundaries.
On the other hand, CAL \cite{rao2021counterfactualattentionlearningfinegrained, chraki2026counterfactual} incorporates counterfactual reasoning into training by verifying that the learned attention is meaningfully better than controlled alternatives. In particular, CIA-MIL \cite{chraki2026counterfactual} introduces a counterfactual causal intervention on attention, evaluating whether the learned attention contributes meaningfully to prediction compared to a controlled random attention. A main drawback of such approaches is the lack of control on the applied perturbations. Indeed, by comparing the learned attention with a random counterpart, they primarily use counterfactual perturbations as a diagnostic tool and do not explicitly structure how evidence should be redistributed across instances during training. To address this, we incorporate CE at training time, and we have the flexibility to learn what perturbations would impact the final prediction.

\section{Methods}
\label{sec:methods}

\subsection{Preliminaries}

\myparagraph{Multiple Instance Learning}In Multiple Instance Learning (MIL), a sample is defined as a set of instances named a bag $
B_i = \{x_{i,1}, \dots, x_{i,N_i}\},
$
where $N_i$ denotes the number of instances in bag $i$. Importantly, in MIL, the supervision is only available at the bag level, i.e. 
%supervision is only available at the bag level. 
% A bag is defined as
each bag is associated with a label $y_i \in \{1,\dots,K\}$ for a $K$-class classification task.
Instance-level labels are unobserved. We focus on embedding-level attention-based MIL models.

\myparagraph{Standard Attention-Based MIL}Each instance $x_{i,j}$ is independently encoded using a feature extractor $\mathcal{E}$ as : $z_{i,j} = \mathcal{E}(x_{i,j}) \in \mathbb{R}^d$. 
\noindent
For clarity, we drop the bag index and denote a bag of embeddings as
$\{z_j\}_{j=1}^{N}$.
\noindent
An attention module $\psi$ produces an attention logit $u_j$ for each instance:
% \begin{equation}
% u_j = \psi(z_j), 
% \qquad 
% a_j = \mathrm{softmax}(u_j).
% \label{eq:abmil_1}
% \end{equation}
\begin{equation}
u_j = \psi(z_j),
\qquad
a = \mathrm{softmax}(u),
\label{eq:abmil_1}
\end{equation}
where $u=(u_1,\ldots,u_N)\in\mathbb{R}^{N}$ denotes the vector of attention logits and
$a=(a_1,\ldots,a_N)\in\mathbb{R}^{N}$ the corresponding normalized attention weights.

\noindent
We denote the bag-level classifier by $\varphi$. The bag representation $\hat{Z}$, class logits $F(u)$ and probabilities $\hat{y}(u)$ are then obtained as:
\begin{equation}
\hat{Z} = \sum_{j=1}^{N} a_j z_j,
\qquad
F(u) = \varphi(\hat{Z}) \in \mathbb{R}^K,
\qquad
\hat{y}(u) = \mathrm{softmax}(F(u)).
\label{eq:abmil_2}
\end{equation}

\noindent

\subsection{Counterfactual Attention Regularization}
\label{sec:cf_attention}

\begin{figure}[t]
    \centering
    \includegraphics[width=1\linewidth]{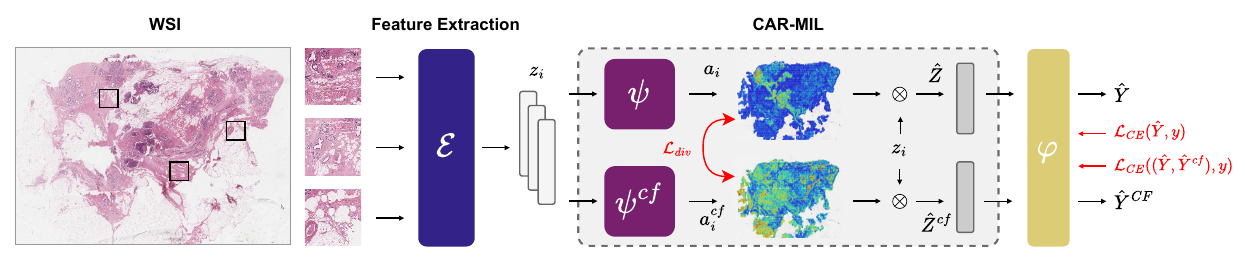}
    \caption{\textbf{\ours Overview.} \ours is a two-branch attention MIL architecture trained with an explainable proximity-evidence constraint.
    Both branches operate on the same instance embeddings and share the downstream classifier. During training, a joint loss encourages
    (i) accurate bag-level predictions from the main(factual) branch,
    (ii) label-consistent evidence separation between main and counterfactual branches via the evidence differential,
    and (iii) meaningful differences between their attention vectors w.r.t the prediction shift.}
    \label{fig:overall}
\end{figure}

\myparagraph{Factual and Counterfactual Branches} Our method follows a two-branch setting (Figure~\ref{fig:overall}) with a factual and counterfactual branches. Both have the same ABMIL~\cite{ilse2018attentionbaseddeepmultipleinstance} architecture, but different weights that are optimized concurrently. The encoder $\mathcal{E}$ and downstream classifier $\varphi$ are not part of the branches and are thus shared between both. The difference between CAR-MIL and ABMIL lies in the additional lightweight counterfactual branch, introducing minimal computational overhead. Differences in predictions thus arise solely from attention weighting differences between the two branches. The attention modules for the factual and counterfactual branches are respectively denoted as $\psi$ and $\psi_{cf}$, and similarly to eq.~\ref{eq:abmil_1}, ~\ref{eq:abmil_2}, we have the following for the counterfactual branch:

\begin{equation}
u^{\mathrm{cf}}_j = \psi_{\mathrm{cf}}(z_j),
\quad
a^{\mathrm{cf}} = \mathrm{softmax}(u^{\mathrm{cf}}),
\quad
\hat{Z}^{\mathrm{cf}} = \sum_{j=1}^{N} a^{\mathrm{cf}}_j z_j,
\quad
F(u^{\mathrm{cf}}) = \varphi(\hat{Z}^{\mathrm{cf}}).
\end{equation}
\noindent

\myparagraph{Evidence Differential }We define the counterfactual evidence differential in logit space as:
\begin{equation}
\Delta F(u,u^{\mathrm{cf}}) 
= 
F(u) - F(u^{\mathrm{cf}}) 
\in \mathbb{R}^K 
\end{equation}
\noindent
It quantifies the difference between the predicted logits of factual and counterfactual attention logits. For a specific class $y\in \{1,\dots,K\}$, $\Delta F_y$ represents the element in $\Delta F$ corresponding to class $y$. A large $\Delta F_y$ indicates that the logit predicted from the factual attention is higher than from its counterfactual counterpart. This would suggest that the factual branch attention provides stronger class-$y$ evidence than the counterfactual branch.

\subsection{Training Objective}

we add to the main classification loss of the model an additional composite loss term constraining the two branches attentions to be close yet their predictions to be different. Therefore, prediction shifts are encouraged to arise from small but structured differences of the attention logits. The proposed \ours framework is therefore optimized using three complementary loss terms:

\begin{itemize}

    \item \textbf{Classification loss:} Standard cross entropy loss on the factual branch prediction. This main loss ensures accurate bag-level prediction from the main (factual) branch:
    \begin{equation}
    \mathcal{L}_{\mathrm{cls}}
    =
    \mathrm{CE}(\hat{y}(u), y)
    \end{equation}
    
    \item \textbf{Counterfactual evidence differential loss: } Encourages the evidence differential 
    $\Delta F$ to favor difference between the two branches predictions on the ground-truth class, enforcing label-consistent separation between factual and counterfactual predictions:
    \begin{equation}
    \mathcal{L}_{\mathrm{diff}}
    =
    \mathrm{CE}\big(\mathrm{softmax}(\Delta F), y\big)
    = 
    \log\!\left(1+\sum_{k\neq y}e^{\Delta F_k-\Delta F_y}\right)
    \end{equation}

     In particular, if $\mathcal{L}_{\mathrm{diff}}\le \varepsilon$, then for all $k\neq y$,
    \begin{equation}
    \Delta F_y-\Delta F_k \ge -\log(e^\varepsilon-1)
    \label{eq:margin_bound}
    \end{equation}
    Thus, minimizing $\mathcal{L}_{\mathrm{diff}}$ explicitly
    encourages the two branches to differ primarily on the ground-truth class.

    \item \textbf{Attention-logit proximity regularization:}  Constrains the counterfactual attention logits to remain close to the factual attention logits, ensuring that prediction differences arise from minimal and controlled weighting updates of attention:
    \begin{equation}
    \mathcal{L}_{\mathrm{div}}
    =
    D(u, u^{\mathrm{cf}}),
    \end{equation}
    where $D(\cdot,\cdot)$ denotes a distance metric in logit space. In practice, $D(\cdot,\cdot)$ can be instantiated using a normalized $L_1$ distance or a cosine-based dissimilarity.  
    The $L_1$ formulation,
    \begin{equation}
    \label{eq:l1_div}
    \mathcal{L}_{\text{div}}
    = D_{L_1}(u, u^{\text{cf}}) = \frac{1}{N}\sum_{j=1}^{N} \big| u_j - u_j^{\text{cf}} \big|,
    \end{equation}
    penalizes absolute deviations in a sparse and interpretable manner, encouraging the counterfactual branch to modify the raw attention values of only a small subset of influential instances.  
    This property aligns with the intuition that counterfactual evidence should emerge from \emph{minimal} attention changes.
    
    Alternatively, a cosine-based distance,
    \begin{equation}
    \label{eq:cos_div}
    \mathcal{L}_{\text{div}}
    = D_{\cos}(u, u^{\text{cf}}) = 1 - 
    \frac{
    \sum_{j=1}^{N} u_j \, u_j^{\text{cf}}
    }{
    \|u\|_2 \, \|u^{\text{cf}}\|_2
    },
    \end{equation}
    captures directional disagreement between the two attention patterns, encouraging the counterfactual head to orient its attention toward different subsets of instances while remaining scale-invariant. This is particularly useful when the magnitude of the logits is less important than their relative orientation across instances.

\end{itemize}

\myparagraph{Overall Objective} The full training objective implements a counterfactual attention regularization principle to guide the learning dynamics of attentions. It encourages solutions where factual and counterfactual attentions remain close but induce label-consistent prediction differences. As a result, the factual attention is encouraged to highlight supporting evidence for the predicted class compared to the counterfactual branch. The final loss is given as follows,

\begin{equation}
\mathcal{L}
=
\mathcal{L}_{\mathrm{cls}}
+
\alpha \mathcal{L}_{\mathrm{diff}}
+
\lambda \mathcal{L}_{\mathrm{div}},
\qquad
\alpha,\lambda \ge 0,
\end{equation}

where $\alpha$ and $\lambda$ are hyperparameters controlling the contribution of each loss in the final objective. \ic{In Sect.\ref{sec:supp_theory} of the supplementary material, we provide additional analytical results that explain why the CAR module increases factual attention on instances supporting the class prediction, while decreasing CF attention on those instances and increasing it on non-supportive ones.}
\section{Experiments and Results}
\label{sec:experiments}

We conducted experiments on both synthetic and histopathological datasets. In the following sections, we summarize the datasets used, along with their implementation details, the comparison methods, and the evaluation strategy.

\subsection{Experiments on MNIST-bags Synthetic Data}

\myparagraph{Datasets} 
We leverage the recent MIL classification datasets from \cite{hense2025xmilinsightfulexplanationsmultiple} derived from MNIST \cite{mnist}. In this synthetic setting, instance-level annotations are available to indicate whether an instance provides supporting or refuting evidence for the bag label according to the dataset construction rule. We consider two dataset variants: \textit{4-bags}: Class 1 if the bag contains an 8, class 2 if it contains a 9, class 3 if it contains both 8 and 9, and class 0 otherwise. This setting evaluates the model’s ability to capture interactions between instances that jointly determine the bag label. \textit{Adjacent Pairs}: a bag is class 1 if it contains any pair of consecutive digits between 0 and 4; otherwise, the bag corresponds to class 0. In this setting, the evidential role of an instance depends on the presence of other instances in the bag, making the prediction inherently contextual. This setup therefore evaluates whether models correctly capture context-dependent evidence relationships between instances.

\myparagraph{Implementation Details}
Feature extraction is performed for the MNIST images using ResNet18 \cite{resnet} model pre-trained on ImageNet \cite{ImageNet} from the TorchVision library \cite{torchvision}. Experiments were repeated 10 times with a learning rate of 0.0002 and the SGD optimizer. We compare our proposed strategy against MIL pooling baselines, reporting means and standard deviations across repetitions of performance in terms of area under the curve (AUC) at the bag level, as well as area under the precision-recall curve of attention prediction of instances' positive evidence label (AUPRC$^+$) and a two-class (positive and negative evidence classes) averaged area under the precision-recall curve (AUPRC$^\pm$). More details are to be found in Sects.\ref{supp_sect_5} and \ref{supp_sect_6} of our supplementary material. We evaluate our method using the $L1$ distance as the metric to measure the similarity between factual and counterfactual attentions. We present the considered baselines along with the method used to extract instance scores in parentheses: ABMIL \cite{ilse2018attentionbaseddeepmultipleinstance} (raw attention scores), TransMIL \cite{shao2021transmiltransformerbasedcorrelated} (attention rollout \cite{abnar2020quantifying}), AddMIL \cite{javed2022additivemilintrinsicallyinterpretable} (patch-level scores).

In all models, the attention is interpreted as representing \textit{positive evidence} (instances that support the predicted label) while reverse attention is used to represent \textit{negative evidence}. In our proposed approach, counterfactual attention serves as an estimation of negative evidence, explicitly modelling instance-level factors that contradict the predicted outcome.

\begin{table}[t]
    \centering
  
        \caption{
    \textbf{Comparison of Baseline and Counterfactual MIL Models on the MNIST-bags Synthetic Datasets.}
    Results are reported for binary (\textit{Adjacent Pairs}) and multi-class (\textit{Four Bags}) classification tasks. 
    \textit{Bag AUC} measures slide-level predictive performance, while \textit{Instance AUPRC$^+$} and \textit{AUPRC$^\pm$} assess alignment between attention scores and instance-level evidence supporting or refuting the bag label. \textit{CAR-MIL} substantially improves instance-level performance, indicating better discriminative evidence of the attention, while maintaining strong bag-level performance.
    }
    \resizebox{1\columnwidth}{!}{%
    \begin{tabular}{lccccccc}

    \toprule
    
    & & \multicolumn{3}{c}{\textbf{Adjacent Pairs}} & \multicolumn{3}{c}{\textbf{Four Bags}}  \\
     \cmidrule(lr){3-5}\cmidrule(lr){6-8}

    \textbf{Model} & \textbf{Evidence} & \textbf{AUC} ($\uparrow$) & \textbf{AUPRC}$^+$ ($\uparrow$) & \textbf{AUPRC}$^\pm$ ($\uparrow$) 
    & \textbf{AUC} ($\uparrow$) & \textbf{AUPRC}$^+$ ($\uparrow$) & \textbf{AUPRC}$^\pm$ ($\uparrow$) 
    \\
    
    \midrule

     ABMIL~\cite{ilse2018attentionbaseddeepmultipleinstance} & attention & 85.7 ± 5.9 & 76.9 ± 6.1 & 61.5 ± 0.9  & 99.1 ± 0.1 & \textbf{86.8 ± 0.2} & 53.1 ± 0.1 \\
    
    AddMIL~\cite{javed2022additivemilintrinsicallyinterpretable} & attention & 81.0 ± 2.2 & 64.3 ± 4.8 & 63.5 ± 1.2  & 97.0 ± 2.7 & 82.1 ± 12.1 & 53.4 ± 0.8  \\
    & patch-scores & N/A & 66.7 ± 7.4 & \underline{67.6 ± 6.6}    
    & N/A & 76.4 ± 15.0 & \underline{76.8 ± 14.6}   \\

    TransMIL~\cite{shao2021transmiltransformerbasedcorrelated} & attention & \textbf{94.6 ± 4.4} & \underline{81.5 ± 6.9} & 61.7 ± 2.1
    & \underline{99.5 ± 0.1} & 81.6 ± 6.3 & 51.9 ± 0.8   \\
    
     \our \ours & \our attention & \our \underline{94.1 ± 5.9} & \our \textbf{85.7 ± 6.5} & \our \textbf{84.6 ± 5.2} & \our \textbf{99.6 ± 0.1} & \our \textbf{86.8 ± 0.6} & \our \textbf{89.7 ± 0.9}  \\
    
    \bottomrule

    \end{tabular}%
    }
    
    \label{tab:synthetic_results}
    
\end{table}

\myparagraph{Counterfactual Reasoning Improves Attention }
The synthetic benchmark provides a controlled setting to evaluate both downstream classification performance and the alignment between attention distributions and instance-level evidence.
For the \textit{Adjacent Pairs} binary classification task, the model must be aware of the contextual influence of individual instances to achieve accurate predictions. As reported in Table \ref{tab:synthetic_results}, CAR-MIL outperforms the ABMIL and AddMIL, and reaches performance comparable to TransMIL at the bag level (94.1 vs.\ 94.6 AUC). More importantly, our method yields substantially higher instance-level performance (AUPRC$^+$=85.7 and AUPRC$^\pm$=84.6), indicating that the learned attention more faithfully reflects discriminative evidence.
As for the \textit{Four Bags} multi-class task, our method improves both performance at the bag-level and attention reliability to reflect the evidence provided by instances that support or refute the bag label. These synthetic experiments suggest that counterfactual reasoning applied at the attention level promotes a tighter correspondence between attention allocation and instance-level evidence. As a result, more faithful attention leads to improved predictive performance.

\subsection{Experiments on Histopathology Data}

\myparagraph{Datasets} We conduct our experiments on multiple public whole-slide image (WSI) datasets from TCGA~\cite{tomczak2015review}: \textit{TCGA-NSCLC} (Non–Small Cell Lung Cancer) and \textit{TCGA-BRCA}(Breast Invasive Carcinoma) for binary cancer subtyping, \textit{TCGA-LUAD}(Lung Adenocarcinoma) for TP53 mutation prediction, \textit{CAMELYON16}~\cite{bejnordi2017diagnostic} for binary metastasis detection in breast lymph node, and \textit{BRACS} \cite{BRACS}(BreAst Cancer Subtyping), a breast cancer dataset for multiclass tissue classification between \textit{normal}, \textit{benign}, and \textit{malignant} categories.. \textit{CAMELYON16} and \textit{BRACS} contain instance-level annotations, with the latter having only sparse labels.

% two different image encoders: \textit{(i)} ResNet50 \cite{resnet} pre-trained on ImageNet \cite{ImageNet} \cite{torchvision} and \textit{(ii)}

\myparagraph{Implementation Details}
Slides were processed using the standard approach as in \cite{clam} to obtain patches of size 256x256 at 20X magnification. We use pre-extracted features with  UNI-V1 \cite{chen2024uni} foundation model. Please note that any other encoder could be integrated into our method. Experiments were conducted under five cross-validation settings for \textit{TCGA-NSCLC}, \textit{TCGA-BRCA}, and \textit{TCGA-LUAD} using a learning rate of 0.0002. We used the originally published train-val-test split for \textit{BRACS} in three runs with a learning rate of 0.0001, and similarly the train-test split for \textit{CAMELYON16} in three runs with a learning rate of 0.0002. All experiments were conducted using the Adam~\cite{adam2014method} optimizer. Additional details to be found in Sects.\ref{supp_sect_5} and \ref{supp_sect_6} of our supplementary material.  

We challenge \ours against several popular MIL frameworks including: baseline approaches that do not use attention (Mean-Max MIL), classical attention-based  MIL (ABMIL \cite{ilse2018attentionbaseddeepmultipleinstance}) 
and attention-focused architectures such as CLAM \cite{clam}, ACMIL \cite{zhang2024attentionchallengingmultipleinstancelearning}, DSMIL \cite{li2021dualstreammultipleinstancelearning}, TransMIL \cite{shao2021transmiltransformerbasedcorrelated}, AddMIL \cite{javed2022additivemilintrinsicallyinterpretable}, and CIA-MIL \cite{chraki2026counterfactual}. We report performance metrics in terms of AUC or balanced accuracy and F1 score. For \textit{CAMELYON16}, we report the AUPRC$^+$, the area under the precision recall curve for attention as a prediction of instance labels. Scores are reported in terms of the average ± standard deviation.

In the proposed framework, the proximity regularization constrains the counterfactual attention logits to remain close to the factual attention logits. We experiment with both $L_1$ and cosine distances as attention similarity measures for our method.
%While the theoretical analysis focuses on the $L_1$ formulation, which induces sparse perturbations in attention space, we also evaluate a cosine-distance variant. 
%Cosine distance constrains the direction of the attention logits while remaining invariant to their magnitude, leading to more distributed and structured reweighting of instances. 
Evaluating both variants allows us to study how different proximity geometries affect the redistribution of evidence and predictive performance.

\setlength\tabcolsep{1.5pt} 

\begin{table*}[t]
    \caption{
    \textbf{Performance Comparison of CAR-MIL Against State-of-the-Art MIL Models on Histopathology Classification Tasks.}
   Results are reported for BRCA and NSCLC cancer subtyping, TP53 mutation prediction in LUAD, BRACS tissue subtyping, and CAMELYON16 metastasis detection. Metrics include AUC, F1-score, and balanced accuracy. \pos{CAR-MIL consistently improves or matches the best baseline performance across datasets using UNI features.}
    }
    \label{tab:model_scores}
    \centering 
    % \scriptsize 
    \resizebox{\textwidth}{!}{ 
    \begin{tabular}{lcccccccccc}
    \toprule
    
    & \multicolumn{2}{c}{\textbf{BRCA}} & \multicolumn{2}{c}{\textbf{NSCLC}} &  \multicolumn{2}{c}{\textbf{LUAD}} & \multicolumn{2}{c}{\textbf{BRACS}} &  \multicolumn{2}{c}{\textbf{CAMELYON16}} \\ \\[-1.05em]

    \cmidrule(lr){2-3} \cmidrule(lr){4-5} \cmidrule(lr){6-7}\cmidrule(lr){8-9}\cmidrule(lr){10-11}
      \textbf{Model} & \textbf{AUC} ($\uparrow$)& \textbf{F1} ($\uparrow$) & \textbf{AUC} ($\uparrow$) & \textbf{F1} ($\uparrow$)  & \textbf{AUC} ($\uparrow$)& \textbf{F1} ($\uparrow$) &  \textbf{BACC} ($\uparrow$)  &  \textbf{F1} ($\uparrow$) &  \textbf{AUC} ($\uparrow$)  &  \textbf{AUPRC$^+$} ($\uparrow$) \\ \\[-1.05em]
    \midrule
    
    Meanmil & 93.2±2.4 & 85.1±2.4  & 96.9±1.3 & 94.0±1.0  & 74.5±6.3 & 71.1±7.4 & 33.2±1.8	& 28.2±2.1 & 62.5±4.8	&  N/A \\ \\[-1.05em]
    
    MaxMIL  & \underline{95.4±1.5} & 88.0±1.1 & 97.5±1.0 & 94.6±1.9   & 76.0±5.4 & 71.8±4.9 &  32.3±7.1 & 29.4±8.6 & 98.3±0.4 & N/A \\ \\[-1.05em]

    CLAM~\cite{clam}  & 94.5±2.3 & 87.6±1.7  & \underline{97.9±0.8} & 94.1±1.7   & 74.0±3.0 & 70.6±1.8 & 40.4±5.5 & 38.0±4.9 & \underline{99.7±0.3} & 94.4±0.3 \\ \\[-1.05em]

    AddMIL~\cite{javed2022additivemilintrinsicallyinterpretable}  & 93.7±2.6 & 86.0±3.3  & 94.6±3.0 & 91.9±2.7  & 73.3±3.9 & 71.5±4.0 & 38.3±9.6 & 35.7±11.6 & 98.2±1.4 & 93.4±1.2 \\ \\[-1.05em]
  
    DSMIL~\cite{li2021dualstreammultipleinstancelearning}  & 94.1±1.6 & 85.3±2.3 & 97.4±1.1 & 94.0±2.6    & 66.9±4.7 & 64.8±5.1 & \underline{42.3±2.0} &	\underline{40.0±2.0} & 98.9±1.1 & 80.6±11.2 \\ \\[-1.05em] 
    
    TransMIL~\cite{shao2021transmiltransformerbasedcorrelated}  & 93.2±2.7 & 87.7±0.8   & 97.8±0.6 & \underline{94.8±2.0}  & 71.7±5.4 & 68.8±4.4 & 40.0±3.6 &  37.6±3.8  & \textbf{99.9±0.1} &  28.2±3.6 \\ \\[-1.05em] 
    ACMIL~\cite{zhang2024attentionchallengingmultipleinstancelearning}  & 94.4±2.9 & 88.7±2.2   & \underline{97.9±0.8} & 93.9±2.5& 75.5±7.4 & \underline{72.0±6.6} &  41.2±1.4 & 38.5±2.4 & 99.4±0.5  & \textbf{95.4±0.5} \\ \\[-1.05em]

    RRT-MIL~\cite{rrt}  & 93.7±1.2 &  84.6±3.4  & 97.3±1.8 & 93.7±2.9  & 75.1±5.2 & 71.8±4.0 & 39.3± 4.8 & 34.5±5.2 & 99.6±0.5 & 94.4±1.1  \\ \\[-1.05em]

    CIA-MIL \cite{chraki2026counterfactual}  &  94.5±1.2 &  87.3±2.7 & 96.5±1.6 &  93.0±2.5 & \underline{77.5±2.5} & 71.1±5.6 & 37.6±2.1 & 35.0±1.6 & 99.2±0.5 & 92.9±1.5 \\ \\[-1.05em]
    
    \midrule
    ABMIL~\cite{ilse2018attentionbaseddeepmultipleinstance} & 95.3±1.7 & \underline{88.9±1.3}   & 97.6±1.0 & 94.2±1.2 & 74.0±4.3 & 72.1±5.4  & 38.2±4.1  & 36.0±4.5  & 98.7±0.3  & 93.2±1.2 \\ \\[-1.05em]
    % \midrule
    \our \ours  (L1) & \our \textbf{95.5±2.2} & \our 87.9±1.5  & \our \textbf{98.0±0.5} & \our 94.4±1.7  & \our 76.4±4.9 & \our \underline{72.4±3.8} & \our 40.8±2.5 & \our 38.1±2.5  & \our \textbf{99.9±0.1} & \our \underline{94.8±1.0} \\ \\
    [-1.05em]
    
    \our \ours (COS) & \our  95.0±1.7 & \our \textbf{89.6±1.4}  & \our 97.5±0.9  & \our \textbf{95.0±1.8}  & \our \textbf{78.1±4.1} & \our \textbf{72.6±4.5}  & \our \textbf{43.1±2.0} & \our \textbf{40.4±2.4}  & \our  99.3±0.7 & \our 91.7±0.6   
    \\ \\[-1.05em]

    \bottomrule
    \end{tabular}
    }
    \label{tab:BRCA-NSCLC}
\end{table*}

\myparagraph{\ours \pos{Achieves Competitive Downstream Performance}}
Table~\ref{tab:BRCA-NSCLC} summarizes the performance of \ours across five histopathology tasks. Overall, \ours achieves performance comparable to or better than strong MIL baselines, \pos{while paired statistical t-tests show that no attention-based MIL method significantly outperforms \ours on any dataset}. Performance gains are more pronounced on the more challenging LUAD and BRACS tasks, suggesting that explicitly guiding attention during training is particularly  beneficial in more challenging settings. In particular, the cosine variant of \ours tends to perform better on the more challenging LUAD and BRACS tasks, while the $L_1$ formulation remains competitive on binary subtyping tasks such as BRCA and NSCLC. On CAMELYON16, where patch-level tumor annotations are well defined and available, \ours maintains competitive patch-level performance (AUPRC$^+$) indicating that redistributing attention through the counterfactual branch does not cause the model to deviate from diagnostically-relevant regions.

% More generally, only 11\% of pairwise comparisons between attention-based MIL methods are statistically significant, highlighting the increasingly saturated performance of current MIL benchmarks

% Consistent improvements are observed over the ABMIL backbone on several tasks. These improvements are more pronounced on datasets where overall performance remains relatively lower, such as LUAD and BRACS, suggesting that explicitly guiding attention during training can be particularly beneficial in more challenging settings.
% %where discriminative signals are weaker. 

\section{\ic{Analysis of Counterfactual Attention Regularization}}
\label{sec:ablation}

\begin{table*}[t]
    \caption{
    \textbf{Ablation with Respect to MIL Settings.}
    Integrating the proposed counterfactual attention regularization (CAR) into more attention-based MIL architectures, including DSMIL, and TransMIL. Performance comparaison is assessed between the original baselines versus their CAR-augmented variants across five datasets. CAR generally improves or maintains performance across most metrics, showing that the proposed regularization can be integrated into diverse MIL attention architectures.
    }
    \label{tab:ablation_attention}
    \centering 
    % \scriptsize 
    \resizebox{\textwidth}{!}{ 
    \begin{tabular}{lcccccccccc}
    \toprule
    
    & \multicolumn{2}{c}{\textbf{BRCA}} & \multicolumn{2}{c}{\textbf{NSCLC}} &  \multicolumn{2}{c}{\textbf{LUAD}} & \multicolumn{2}{c}{\textbf{BRACS}} &  \multicolumn{2}{c}{\textbf{CAMELYON16}} \\ \\[-1.05em]

    \cmidrule(lr){2-3} \cmidrule(lr){4-5} \cmidrule(lr){6-7}\cmidrule(lr){8-9}\cmidrule(lr){10-11}
      \textbf{Model} & \textbf{AUC} ($\uparrow$)& \textbf{F1} ($\uparrow$) & \textbf{AUC} ($\uparrow$) & \textbf{F1} ($\uparrow$)  & \textbf{AUC} ($\uparrow$)& \textbf{F1} ($\uparrow$) &  \textbf{BACC} ($\uparrow$)  &  \textbf{F1} ($\uparrow$) &  \textbf{AUC} ($\uparrow$)  &  \textbf{AUPRC$^+$} ($\uparrow$) \\ \\[-1.05em]
    \midrule
    
    ABMIL~\cite{ilse2018attentionbaseddeepmultipleinstance}  & \textbf{95.3±1.7} & 88.9±1.3   & 97.6±1.0 & 94.2±1.2 & 74.0±4.3 & 72.1±5.4  & 38.2±4.1  & 36.0±4.5  & 98.7±0.3  & 93.2±1.2 \\ \\[-1.05em]
    % \midrule
    \our \ours  & \our 95.0±1.7 & \our \textbf{89.6±1.4}  & \our \textbf{98.0±0.5} & \our 94.4±1.7  & \our \textbf{78.1±4.1} & \our \textbf{72.6±4.5}& \our \textbf{43.1±2.0} & \our \textbf{40.4±2.4}& \our \textbf{99.9±0.1} & \our \textbf{94.8±1.0} \\ \\
    [-1.05em]
    
    \midrule
    DSMIL~\cite{li2021dualstreammultipleinstancelearning}  & 94.1±1.6 & 85.3±2.3 & 97.4±1.1 & 94.0±2.6    & 66.9±4.7 & 64.8±5.1 & \underline{42.3±2.0} &	\underline{40.0±2.0} & 98.9±1.1 & 80.6±11.2 \\ \\[-1.05em] 
    
    \our CAR-DSMIL & \our 94.5±2.2  & \our 85.6±2.7   & \our 97.8±1.0  & \our 94.0±2.3   & \our 71.2±4.1 & \our 67.8±3.4 & \our 42.5±5.1 & \our 39.8±2.0 & \our 99.0±0.6 & \our 87.8±2.9  \\ \\[-1.05em]
    
    \midrule
    TransMIL~\cite{shao2021transmiltransformerbasedcorrelated}  & 93.2±2.7 & 87.7±0.8   & 97.8±0.6 & \textbf{94.8±2.0}  & 71.7±5.4 & 68.8±4.4 & 40.0±3.6 &  37.6±3.8  & \textbf{99.9±0.1} &  28.2±3.6 \\ \\[-1.05em] 
    
    \our CAR-TransMIL  & \our 94.0±1.3 & \our 87.2±4.3  & \our 97.6±1.4 & \our 94.3±2.0 & \our 73.6±4.4 & \our 69.8±4.0 & \our 42.6±6.1 & \our 39.4±6.2 & \our 99.7±0.1 &  \our 32.6±0.3 \\ \\[-1.05em]

    \bottomrule
    \end{tabular}
    }
\end{table*}

\begin{figure}[tb]
    \begin{minipage}[b]{.62\linewidth}
  \centering
    \includegraphics[width=\linewidth]{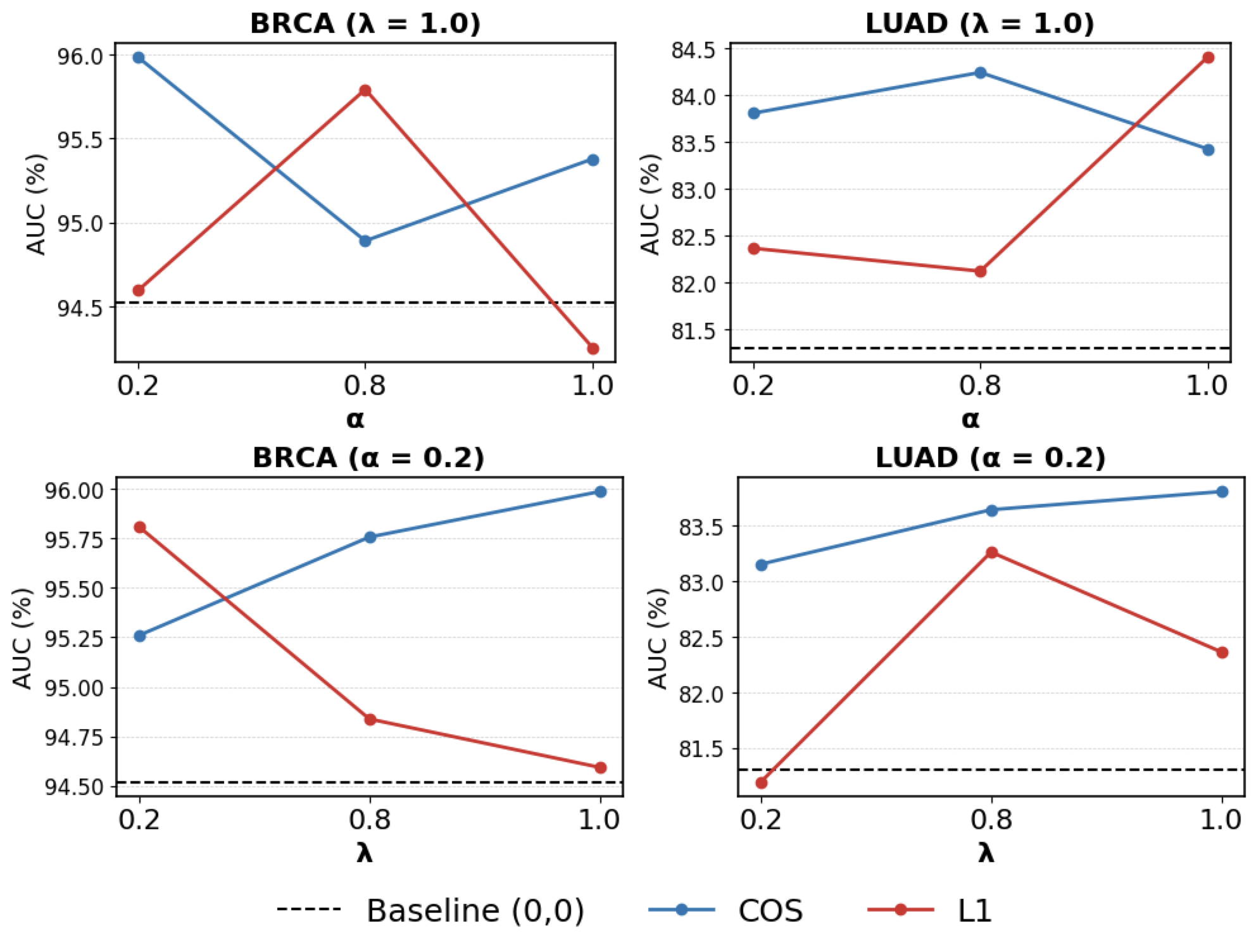}
    \caption{\textbf{Hyperparameters Sensitivity.} Sensitivity to hyperparameters $\alpha$ and $\lambda$ is evaluated on BRCA and LUAD for the Cosine and $L_1$ divergence variants on a single run. Cosine variant tends to have more consistent trends than the L1 variant. }
    \label{fig:short-a}
    \end{minipage} \hfill
     \begin{minipage}[b]{.35\linewidth}
     % \begin{table*}[t]
    \setlength{\tabcolsep}{1pt}
    \captionof{table}{
    \textbf{Ablation on the Feature Encoder Using ResNet50 Features}. Comparison between ABMIL and \ours across three TCGA datasets -- BR.: BRCA, NS.: NSCLC, LU.: LUAD. \ours consistently improves AUC and F1 over the baseline ABMIL indicating that the proposed CAR remains effective even when using standard convolutional features.}
    \label{tab:ablation_features}
    \centering 
    {
    \scriptsize
    % \resizebox{0.7\textwidth}{!}{ %

    \begin{tabular}{llcc}
    \toprule
    
      & \textbf{Model} & \textbf{AUC} ($\uparrow$)& \textbf{F1} ($\uparrow$)  \\
    \midrule
    \multirow{2}{*}{\rotatebox{90}{\tiny \textbf{BR.}}} & ABMIL~\cite{ilse2018attentionbaseddeepmultipleinstance}  & 89.2±2.6 & 79.7±3.9 \\
    & \our \ours  & \our \textbf{90.3±2.1} & \our \textbf{81.9±5.0} \\
    \midrule
    \multirow{2}{*}{\rotatebox{90}{\tiny \textbf{NS.}}} & ABMIL~\cite{ilse2018attentionbaseddeepmultipleinstance}  & 93.4±1.7&  88.2±1.6 \\
    & \our \ours  & \our \textbf{94.3±1.2}  & \our \textbf{88.9±1.3}  \\
    \midrule
    \multirow{2}{*}{\rotatebox{90}{\tiny \textbf{LU.}}} & ABMIL~\cite{ilse2018attentionbaseddeepmultipleinstance} & 68.2±5.6 & 66.3±4.8\\ 
    & \our \ours  & \our \textbf{70.4±6.1}  & \our \textbf{68.4±5.1}  \\
  
    \bottomrule
    
    % \begin{tabular}{lcccccc}
    % \toprule
    
    % & \multicolumn{2}{c}{\textbf{BRCA}} & \multicolumn{2}{c}{\textbf{NSCLC}} &  \multicolumn{2}{c}{\textbf{LUAD}} \\ \\[-1.05em]
    
    % \cmidrule(lr){2-3} \cmidrule(lr){4-5} \cmidrule(lr){6-7}\cmidrule(lr){8-9}\cmidrule(lr){10-11}
    %   \textbf{Model} & \textbf{AUC} ($\uparrow$)& \textbf{F1} ($\uparrow$) & \textbf{AUC} ($\uparrow$) & \textbf{F1} ($\uparrow$)  & \textbf{AUC} ($\uparrow$)& \textbf{F1} ($\uparrow$)  \\ \\[-1.05em]
    % \midrule
    
    % ABMIL~\cite{ilse2018attentionbaseddeepmultipleinstance}  & 89.2±2.6 & 79.7±3.9   & 93.4±1.7&  88.2±1.6 & 68.2±5.6 & 66.3±4.8\\ \\[-1.05em]
    % % \midrule
    % \our \ours  & \our 90.3±2.1 & \our 81.9±5.0  & \our 94.3±1.2  & \our 88.9±1.3  & \our 70.4±6.1  & \our 68.4±5.1  \\ \\
    % [-1.05em]
  
    % \bottomrule
    \end{tabular}
    }
% \end{table*}
     \end{minipage}
\end{figure}

\myparagraph{Hyperparameter Sensitivity }
\cref{fig:short-a} analyzes the sensitivity to hyperparameters $\alpha$ controlling the evidence differential loss $\mathcal{L}_{\mathrm{diff}}$,  and $\lambda$ controlling the attention logit proximity loss $\mathcal{L}_{\mathrm{div}}$. We report the AUC obtained for both the cosine and $L_1$ variants of a run of \ours on the BRCA and LUAD datasets for different combinations of these weights. The explored configurations include values $\{0.2, 0.8, 1.0\}$, together with the baseline ABMIL setting without counterfactual regularization ($\alpha=\lambda=0$). On BRCA, the baseline AUC of $94.5$ improves to values between $95.3$ and $96.0$ depending on the configuration, with the best performance obtained for moderate counterfactual regularization. On LUAD, where the task is more challenging, the baseline AUC of $81.3$ increases to values up to $84.2$--$84.4$. While improvements are observed on both datasets, the effect of the hyperparameter choices is more visible on LUAD than on BRCA. This difference likely reflects the relative difficulty of the tasks: when baseline performance is already high, as on BRCA, performance differences across configurations remain relatively small, whereas on the more challenging LUAD task the influence of the regularization weights becomes more visible. Across settings, the cosine variant exhibits smoother trends across configurations, while the L1 variant shows slightly larger variability depending on the choice of weights.

\begin{figure}[t]
    \centering
    \begin{minipage}[b]{.55\linewidth}
    \includegraphics[width=\linewidth]{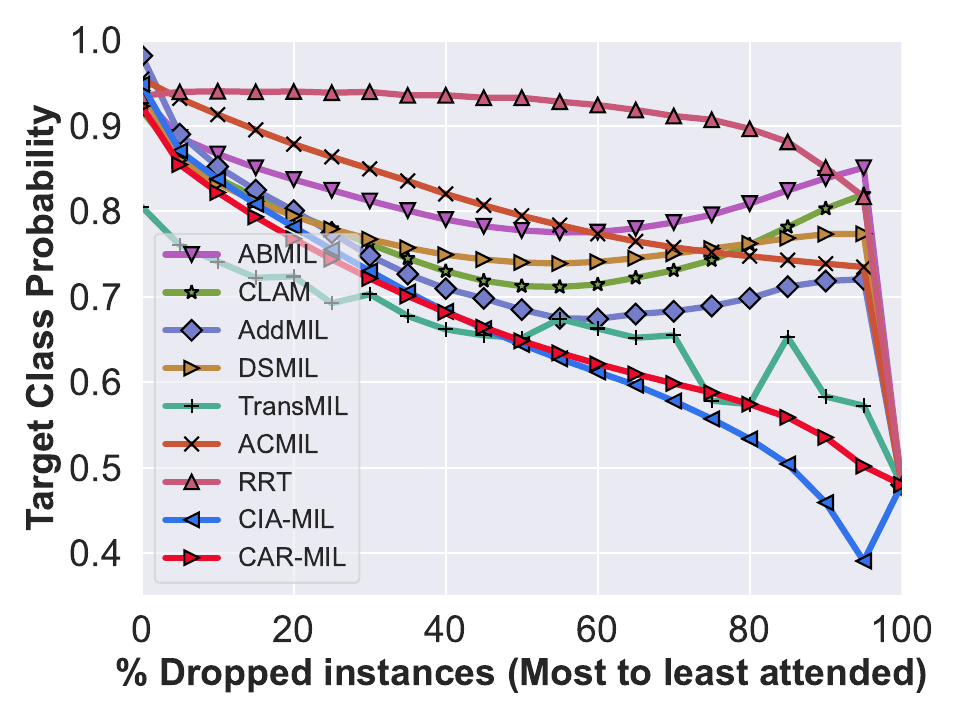}
    \caption{ \textbf{Effect of Removing Highly Attended Instances on the Prediction Confidence.} Removing attention according to top-ranked instances shows that \ours yields a smooth, monotonic drop in confidence, indicating faithful instance importance, whereas ABMIL exhibits non-monotonic behavior.  }
    \label{fig:flap}
    \end{minipage}
    \begin{minipage}[b]{.4\linewidth}
    % \begin{table}[ht]
    \setlength{\tabcolsep}{1pt}
    \centering
    {
    \tiny
    \captionof{table}{\textbf{Attention as Interpretability Proxy on MNIST-bags.} Evaluation of attention as a proxy for instance-level evidence on synthetic MIL datasets with ground-truth evidence labels. CAR-MIL attention is reliable as an interpretability signal, outperforming ABMIL raw attention and perturbation evidence attribution for explaining ABMIL. -- Rand.: Random, Att.: Attention, Pert.: Perturbation, Adj. P.: Adjacent Pairs, Four B: Four Bags.}
    \label{tab:interp_att}
    % \resizebox{0.7\textwidth}{!}{%

    \begin{tabular}{lllcc}
    \toprule
    & \textbf{Model} & \textbf{Evid.} & \textbf{AUPRC}$^+$) & \textbf{AUPRC}$^\pm$ \\
    \midrule
    
    \multirow{4}{*}{\rotatebox{90}{\textbf{Adj. P.}}} & & Rand. & 54.2 ± 0.9 & 54.2 ± 0.9 \\
    & ABMIL~\cite{ilse2018attentionbaseddeepmultipleinstance} & Att. & 76.9 ± 6.1 & 61.5 ± 0.9 \\
    && Pert. & \underline{78.5 ± 1.0} & \underline{78.5 ± 1.0} \\
    & \our  \ours & \our Att. &  \our \textbf{85.7 ± 6.5} & \our \textbf{84.6 ± 5.2} \\
    \midrule

    \multirow{4}{*}{\rotatebox{90}{\textbf{Four B.}}} & & Rand. & 30.6 ± 0.2 & 31.2 ± 0.2 \\
    & ABMIL~\cite{ilse2018attentionbaseddeepmultipleinstance} & Att. & \textbf{86.8 ± 0.2}  & 53.1 ± 0.1 \\
    && Pert. & \underline{87.9 ± 1.6} & \underline{87.7 ± 2.1} \\
    & \our  \ours & \our Att. & \our \textbf{86.8 ± 0.6}  & \our \textbf{89.7 ± 0.9} \\
    
    \bottomrule

    \end{tabular}
    }
% \end{table}
    \end{minipage}
\end{figure}

\begin{figure}[t]
    \centering
    \includegraphics[width=1\linewidth]{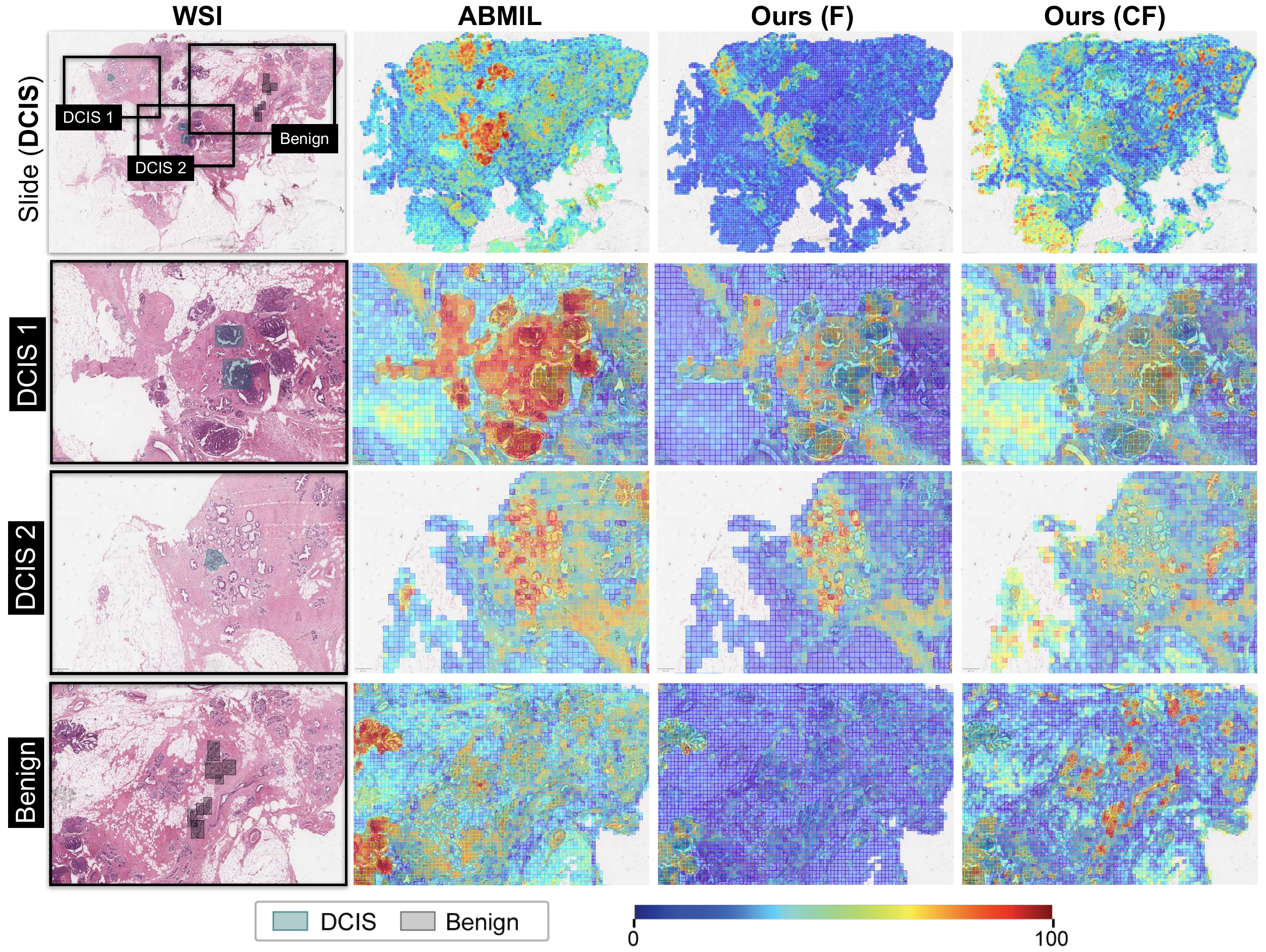}
    \caption{\textbf{Comparisons Between the Attentions of ABMIL and the Two \ours Attention Branches (F: Factual, CF: Counterfactual).} Qualitative comparison of attention maps on a DCIS WSI from the BRACS dataset. ABMIL highlights broad regions within the annotated DCIS area but produces a diffuse attention pattern. Our model (factual attention) concentrates attention on more precise subregions that correspond closely to the annotated malignant ducts. The counterfactual branch highlights benign or non-malignant areas, providing a complementary view of “negative evidence”. Together, the two branches provide a more structured and interpretable separation between supporting and refuting regions.}
    \label{fig:placeholder}
\end{figure}

\myparagraph{MIL Setting }
Table~\ref{tab:ablation_attention} evaluates the effect of integrating our counterfactual attention regularization (CAR) into different MIL settings: ABMIL \cite{ilse2018attentionbaseddeepmultipleinstance} (already presented in Table~\ref{tab:model_scores}, but reported for comparison with other MIL methods), DSMIL\cite{li2021dualstreammultipleinstancelearning} and TransMIL\cite{shao2021transmiltransformerbasedcorrelated}. We provide in Sect.\ref{supp_sect_7} of our supplementary material more technical details on the integration of CAR following the different attention formulations. For each setting, we report the best-performing distance variant (either $L_1$ or cosine) in order to focus on the contribution of the proposed CAR mechanism rather than the choice of distance metric. Across architectures, the proposed regularization consistently improves or maintains performance compared to the original models. In particular, the gains are most visible for the ABMIL backbone. Similar trends are observed when integrating the method into DSMIL and TransMIL, where CAR variants improve mutation prediction LUAD and BRACS classification performance while preserving strong results on easier tasks. Overall, these results indicate that counterfactual attention regularization can be effectively combined with different MIL aggregation mechanisms to improve slide-level prediction.

\myparagraph{Feature Extractor }
Table~\ref{tab:ablation_features} reports an analysis with respect to the feature encoder. Using  pretrained ResNet50 \cite{resnet} on ImageNet \cite{ImageNet,torchvision} features, \ours consistently improves performance over the ABMIL baseline across all datasets. As expected, absolute performance remains lower than with UNI features reported in the main experiments, reflecting the stronger representational capacity of pathology-specific foundation models. Nevertheless, the consistent gains demonstrate that the proposed counterfactual attention regularization is not tied to a specific encoder and can improve MIL aggregation even with standard convolutional features. In fact the gains are even more pronounced with out-of-domain features as the ones from ResNet50 compared to UNI features, highlighting the positive impact of guiding the learning dynamics of attention under generic features representations.

\ic{\myparagraph{Multi-class Setting } We investigate the counterfactual branch predicted class logit under the multi-class setting. We find that the class whose CF logit increases most varies with the ground-truth class. On Four Bags, class 1 shifts toward class 3 (100\% of bags) while class 3 shifts toward class 1 (62\%), reflecting the dataset evidence structure. On BRACS, redistribution is diffuse across all 7 classes with no dominant competitor, consistent with higher semantic ambiguity.}
\section{Attention Analysis}
\label{sec:att_analysis}

\myparagraph{Counterfactual Attention Regularization for Interpretability} We assess how much the proposed counterfactual attention regularization contributes to bridging the gap between predictive performance and interpretability relevance in MIL frameworks. 
Using the synthetic benchmark, we compare the attention from our method with the raw attention from \textit{ABMIL} and attention maps generated from two baselines: 
(i) a \textit{Random} assignment of attention weights to instances, serving as a lower bound, and 
(ii) a post-hoc explainability method, \textit{Perturbation Single} \cite{early2022modelagnosticinterpretabilitymultiple,hense2025xmilinsightfulexplanationsmultiple}, which estimates instance importance by measuring prediction sensitivity to perturbations, by passing bags of single instances through the model ("single").
We then assess the interpretability of the inherent learned attention in \ours counterfactual variant.  
As shown in Table~\ref{tab:interp_att}, counterfactual-guided attention learning substantially improves the alignment between attention and instance-level evidence compared to both ABMIL and the perturbation-based explanation. In particular, \ours achieves the highest AUPRC$^+$ and AUPRC$^\pm$ scores on both tasks. \\

\myparagraph{Faithfulness Evaluation via Perturbation Analysis}
To further assess the faithfulness of attention, we perform a MORF (Most Relevant First) perturbation test \cite{hense2025xmilinsightfulexplanationsmultiple,early2024inherentlyinterpretabletimeseries}, where instances in correctly-predicted bags are removed in order of decreasing attention importance. For \ours, importance is computed from the discrepancy between factual and counterfactual attention scores. A faithful attention mechanism is expected to produce a \textit{monotonic decrease} in the model’s confidence, since removing the most relevant instances should consistently degrade predictive performance.
In Figure~\ref{fig:flap}, our proposed \ours model exhibits a smooth and monotonic drop in prediction probability as a function of the percentage of removed patches, suggesting that learned attentions align closely with true predictive importance.
In contrast, models such as ABMIL show non-monotonic or convex response curves, suggesting that their attention weights may not reliably capture relevant evidence among instances. Interestingly, CAR-MIL showcases a similar trend as the recent CIA-MIL method that focuses on causally aligning attention with predictions at the cost of a trade-off between performance and explainability. As shown in Table~\ref{tab:model_scores}, \ours outperforms CIA-MIL in terms of downstream performance. Indeed, by improving solely explainability through counterfactual intervention, CIA-MIL can lead to a drop in performance compared with CAR-MIL.

\myparagraph{WSI Heatmaps} To qualitatively illustrate the differences between attentions, we visualize attention maps for a representative WSI from the \textit{BRACS} dataset labeled as \textit{DCIS} (\textit{Ductal Carcinoma In Situ}). Figure~\ref{fig:placeholder} displays the attentions obtained with ABMIL, \ours factual attention (F), and \ours counterfactual (CF) branch. Although ABMIL broadly highlights areas containing annotated DCIS regions, its attention is spatially diffuse. In contrast, \ours produces more localized and concentrated attention regions. The counterfactual head provides further informative signal: its highest value occurs in benign or non-neoplastic areas, \textit{i.e.,} regions whose removal would weaken the model’s confidence in the positive class. This complementary pattern aligns with our design goal: factual attention captures supporting evidence, while counterfactual attention highlights refuting or neutral evidence. \ic{More quantitatively, the factual branch does not simply reproduce ABMIL localization (Spearman $r{=}0.67$ on the representative example in Figure~\ref{fig:placeholder} between attention vectors), while the factual and counterfactual branches attend to distinct regions ($r{=}0.29$). Across BRACS slides, factual attention remains only moderately correlated with ABMIL (L1: $r{=}{0.69}{\pm}0.05$; Cosine: $r{=}{0.69}{\pm}0.03$), whereas factual and counterfactual attentions are strongly anti-correlated under L1 ($r{=}{-}0.67{\pm}0.02$) and weakly correlated under cosine ($r{=}{0.20}{\pm}0.63$). Our supplementary material Figs.\ref{fig:placeholder_supp}, \ref{fig:comparaison} further illustrate this behavior and show that \ours concentrates attention on DCIS regions while suppressing benign tissue compared with ABMIL, DSMIL, AddMIL, and CLAM.}

\section{Conclusion}
\label{sec:conclusion}

In this work, we introduce \ours, a counterfactual attention regularization framework designed to jointly improve the performance and reliability of attention-based MIL models. Motivated by the observation that attention plays a dual role, as both the mechanism that aggregates instance information and the most commonly used proxy for MIL interpretability, we propose a double-branch (factual-counterfactual) architecture and a training strategy that explicitly guides attention using learned counterfactual perturbations. By encouraging the counterfactual branch to induce prediction changes while remaining close to the main attention distribution, \ours reveals informative regions that standard attention may overlook. Through experiments on synthetic datasets and multiple whole-slide image benchmarks, \pos{we demonstrate that \ours maintains or improves downstream performance} while producing attention maps that better align with instance-level evidence. These results suggest that integrating counterfactual explainability reasoning into the attention optimization process offers a promising direction for designing MIL models that are not only more accurate but also more interpretable. Future work could explore how the learned counterfactual attentions can be leveraged beyond training, for instance as auxiliary supervisory signals or for downstream tasks such as improving weak supervision, guiding active learning strategies, supporting model debugging, and facilitating deeper analysis of model behavior and decision-making processes.

% The paper ends with a conclusion. 

% \clearpage\mbox{}Page \thepage\ of the manuscript.
% \clearpage\mbox{}Page \thepage\ of the manuscript.
% \clearpage\mbox{}Page \thepage\ of the manuscript.
% \clearpage\mbox{}Page \thepage\ of the manuscript.
% \clearpage\mbox{}Page \thepage\ of the manuscript. This is the last page.
% \par\vfill\par
% Now we have reached the maximum length of an ECCV \ECCVyear{} submission (excluding references and acknowledgements).
% References should start immediately after the main text, but can continue past p.\ 14 if needed. 
% \clearpage  % TODO FINAL: This \clearpage needs to be removed from both review and camera-ready versions.

\section*{Acknowledgements}
This work has benefited from state financial aid, managed by the Agence Nationale de Recherche under the investment program integrated into France 2030, project references ANR-21-RHUS-0003, ANR-21-CE45-0007, ANR-23-CE45-0029, ANR-23-IAHU-0002, and ANR-23-IACL-0003 – DATAIA CLUSTER (as part of IA CLUSTER program). This project has partily received funding from the European Union’s Horizon Europe research and innovation programme under grant agreement No 101156771. Views and opinions expressed are however those of the authors only and do not necessarily reflect those of the European Union. The European Union cannot be held responsible for them. Experiments have been conducted using HPC resources from the Mésocentre computing center of CentraleSupélec and École Normale Supérieure Paris-Saclay, supported by CNRS and Région Île-de-France, and resources from GENCI–IDRIS (Grant 2025-AD011015828, 2026-AD011015828R1). The results shown in this paper are part based upon data generated by the TCGA Research Network: https://www.cancer.gov/tcga.
% ---- Bibliography ----
%
% BibTeX users should specify bibliography style 'splncs04'.
% References will then be sorted and formatted in the correct style.
%

\bibliographystyle{splncs04}
\bibliography{main}

\newpage

\appendix
\renewcommand{\thesection}{A.\arabic{section}}
% Continuous numbering across the whole supplementary, prefixed with "A."
\renewcommand{\thefigure}{A.\arabic{figure}}
\renewcommand{\thetable}{A.\arabic{table}}

% Reset counters to 0 so figures/tables in the appendix start fresh at A.1
% (remove these two lines if you want them to continue from the main paper's numbering)
\setcounter{figure}{0}
\setcounter{table}{0}

\begin{center}
{\Large \textbf{Supplementary Material}}
\end{center}

% \section*{Supplementary Material}

\section{Background on Counterfactual Explanations}

Counterfactual explanations are a conceptual framework for machine learning interpretability. According to \cite{molnar2020general}, \textit{a counterfactual explanation of a prediction describes the smallest change to the input values that shifts the prediction to a predefined output}. This type of reasoning has its roots in social sciences, describing individuals imagining hypothetical scenarios contradicting factual reality, that would have changed the outcome of a certain situation. \cite{molnar2020general, wachter2018counterfactualexplanationsopeningblack}. 

Searching for a counterfactual close enough to the factual inputs but yielding different outcomes can result in many possible options. According to \cite{molnar2020general}, the counterfactual choice should take into consideration its likelihood in real world. However in our work we operate at the level of attention over bags' instances. Given that there is no access to a ground truth of a likely attention distribution, our counterfactual regularization mainly builds upon the Wachter et al \cite{wachter2018counterfactualexplanationsopeningblack}  formulation of the counterfactual cost function as it was first introduced for interpretability. Given a sample $\mathbf{x}$, the goal is to find a counterfactual $\mathbf{x^{\text{cf}}}$ whose prediction is close to a desired target $y^{\text{cf}}$ while staying as close as possible to the original input. Wachter et al.\ propose minimizing the following cost function:

\begin{equation}
\label{eq:wachter_loss}
\mathcal{L}(\mathbf{x},\mathbf{x^{\text{cf}}}, y^{\text{cf}}, \lambda)
= \alpha \cdot \ell\big(f(\mathbf{x^{\text{cf}}}),\, y^{\text{cf}}\big)
+ d(\mathbf{x},\mathbf{x^{\text{cf}}})
\end{equation}

where $f(\cdot)$ is the prediction model, $\ell(\cdot)$ is a loss function penalizing deviation from the desired output $y^{\text{cf}}$, and $d(\mathbf{x},\mathbf{x^{\text{cf}}})$ is the distance between the counterfactual and the original input. In this work, instead of operating in the input space (e.g.\ image pixels), we target the attention module.

\section{Comparison with Weakly Supervised Knowledge Distillation}

WENO~\cite{qu2022bidirectionalweaklysupervisedknowledge} proposes a weakly supervised knowledge distillation framework that combines an attention-based MIL bag classifier with an instance classifier. In this approach, attention scores produced by the bag classifier are interpreted as soft pseudo-labels used to supervise the instance classifier, which in turn performs hard positive instance mining to refine the bag classifier.

Although WENO and our approach both rely on attention-based MIL architectures, their objectives differ. WENO assumes that attention scores can approximate class-specific supervision at the instance level, effectively learning patch-level predictions from slide-level labels in digital histopathology datasets. This assumption is well aligned with binary MIL tasks such as CAMELYON16 metastasis detection, where positive regions correspond to localized tumor areas that can be reasonably associated with instance-level labels.

In contrast, our framework does not attempt to assign class labels to individual instances. Instead, CAR-MIL analyzes how instances support or refute the bag-level prediction through counterfactual attention contrast. This design avoids tying evidence to explicit patch-level class assignments and therefore generalizes naturally to multi-class classification tasks, where defining instance-level labels as positive or negative becomes more ambiguous.

\begin{table}[t]
\centering
\caption{Comparison between WENO and CAR-MIL on representative datasets.}
\label{tab:weno_comparison}
\begin{tabular}{lcccc}
\toprule
& \multicolumn{2}{c}{\textbf{LUAD}} & \multicolumn{2}{c}{\textbf{CAMELYON16}} \\
\cmidrule(lr){2-3} \cmidrule(lr){4-5}
\textbf{Method} & \textbf{AUC} ($\uparrow$) & \textbf{F1} ($\uparrow$) & \textbf{AUC} ($\uparrow$) & \textbf{AUPRC$^+$} ($\uparrow$) \\
\midrule
ABMIL~\cite{ilse2018attentionbaseddeepmultipleinstance} & 74.0±4.3 & 72.1±5.4 & 98.7±0.3 & 93.2±1.2 \\
WENO~\cite{qu2022bidirectionalweaklysupervisedknowledge} & 55.5±10.7 & 58.6±8.6 & 99.3±0.5 & 90.6±2.9 \\
CAR-MIL (ours, L1) & 76.4±4.9 & 72.4±3.8 & 99.9±0.1 & 94.8±1.0 \\
\bottomrule
\end{tabular}
\end{table}

To provide a representative comparison, we evaluated WENO on two datasets used in our experiments. We report AUC and F1 for LUAD and AUC together with AUPRC$^+$ for CAMELYON16.
While WENO performs strongly on the binary CAMELYON16 metastasis detection task, its performance on LUAD mutation prediction is substantially lower. Mutation prediction tasks typically rely on subtle morphological signals that may be spatially diffuse across the slide rather than localized to specific patches. In such settings, enforcing patch-level pseudo-labels can introduce noisy supervision. By contrast, CAR-MIL models prediction evidence through factual vs counterfactual attention allocation without requiring explicit patch-level supervision, which leads to more stable performance across tasks.

\section{Visualisations and Attention Study}

\begin{figure}[t]
    \centering
    \includegraphics[width=1\linewidth]{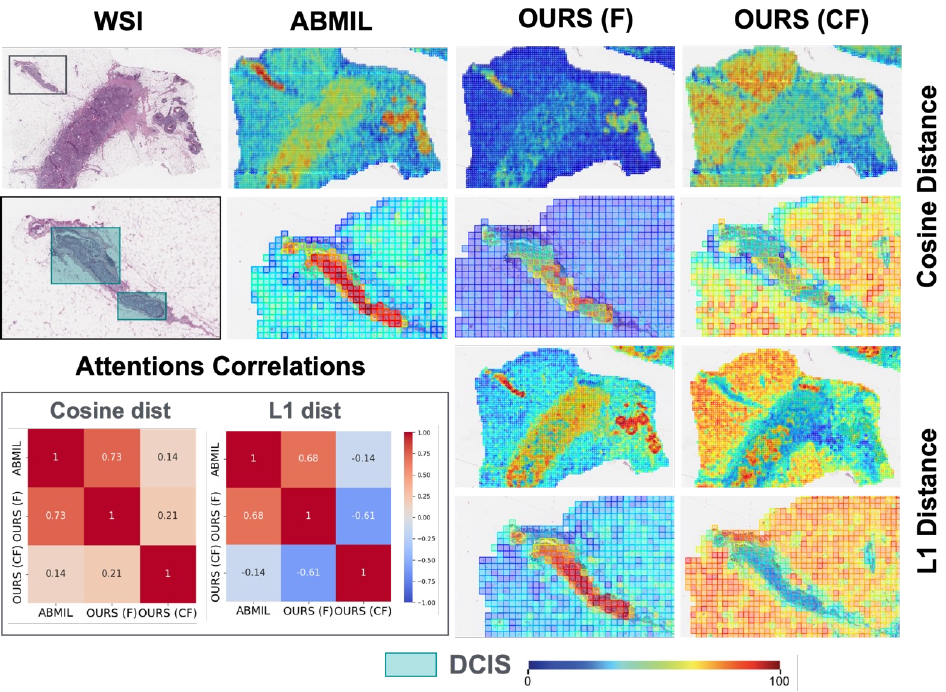}
    \caption{\textbf{Comparison of attention maps across ABMIL and \ours under $L_1$ and cosine divergences.}  Right Bottom: Pearson correlation heatmaps measuring similarity between ABMIL, F, and CF attention maps for each attention distance metric choice for \ours. Under both $L_1$ and cosine, the F head resembles ABMIL closely, while CF exhibits inverse behaviour. Under cosine dissimilarity, F is more concentrated and sharply focused on DCIS regions. All attention maps use Min-Max scaling applied to unormalised attentions.}

    \label{fig:placeholder_supp}
\end{figure}

\myparagraph{Attention Distance Metric Choice}To better understand the effect of the distance metric choice on the behaviour of the main (F) and 
counterfactual (CF) attention branches, we perform a qualitative analysis on a representative BRACS slide containing ductal carcinoma in situ (DCIS) regions annotated by expert pathologists.  
For this slide, we compare the attention maps (F and CF) produced by \ours under $L_1$ 
setting, and cosine setting, together with the baseline ABMIL attention map. For consistency, all attention maps are derived from the unormalised attention scores , 
followed by Min–Max scaling to the range $[0,1]$.

We additionally extract a high-resolution zoom around the annotated DCIS region to examine how the different attention mechanisms highlight diagnostically relevant structures. As shown in figure \ref{fig:placeholder_supp}, across both distance metrics choices, the F and CF attentions exhibit complementary behaviour with main factual attention focusing on regions that support the predicted bag label, while the counterfactual attention tends to emphasise areas deviating from these regions. In particular, under $L_1$ distance, the factual attention (F) most closely resembles the ABMIL attention, producing smooth attention that extend into the surrounding tissue. In contrast, the cosine-based model produces a more concentrated fatual attention, with sharper and more  localized focus around DCIS structures and reduced spread into neighbouring regions.  
The counterfactual attention display the opposite trend in each case. 

\begin{figure}[t]
    \centering
    \includegraphics[width=0.7\linewidth]{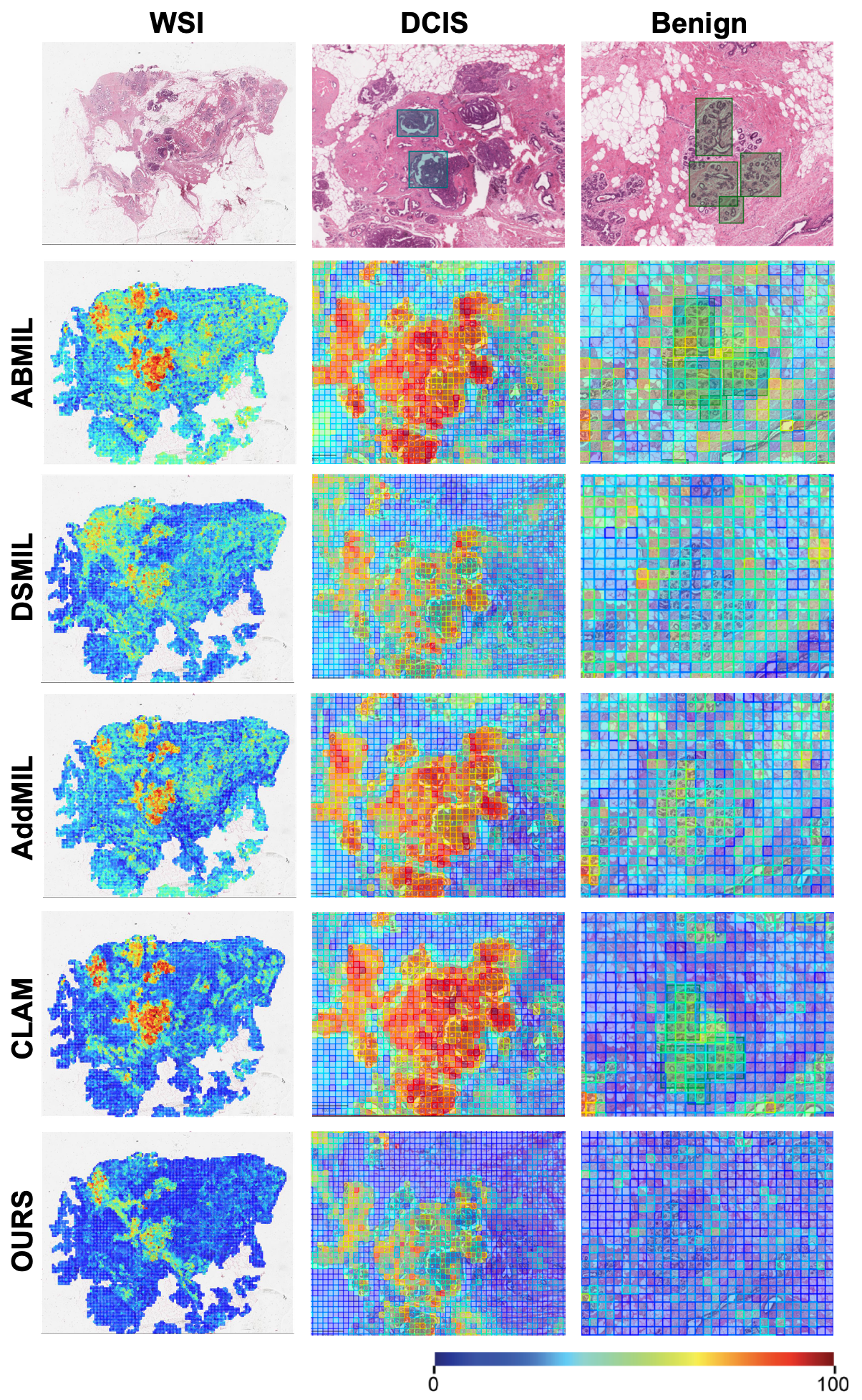}
    \caption{
\textbf{Comparison of attention maps across MIL baselines (ABMIL, CLAM, AddMIL, DSMIL) and 
\ours (F).}  
For a BRACS WSI labelled as DCIS, we compare overall WSI attention maps and two zoomed 
regions: an expert-annotated DCIS region (middle) and a benign region (right).  
Baseline methods tend to distribute attention broadly and often allocate non-negligible weight to 
benign structures.  
In contrast, \ours produces a more concentrated and selective attention pattern, focusing sharply on 
the DCIS lesion while suppressing attention in benign tissue.    
All attention maps are derived from Min–Max scaled unnormalised attentions}

    \label{fig:comparaison}
\end{figure}

\myparagraph{Baselines Comparison} We further compare the behaviour of our main attention (F) with several MIL 
baselines, including ABMIL, CLAM, AddMIL, and DSMIL.  
Figure~\ref{fig:comparaison} shows WSI attention maps for a representative BRACS slide 
labelled as DCIS, together with two high-resolution zooms: one centred on the annotated DCIS region 
and another on a benign area within the same slide.

\section{Theoretical Analysis}
\label{sec:supp_theory}

This section provides additional theoretical analysis of the proposed counterfactual attention regularization framework. We adopt the same notation as in the main paper, and references to equations numbered in the main paper are explicitly indicated.

Intuitively, the proposed regularization encourages the factual and counterfactual attention branches to produce different prediction scores while remaining close in attention-logit space. The analysis below studies how small perturbations of the attention logits affect the relative prediction scores of different classes.
Our goal is therefore to analyze how perturbations of the attention logits influence the counterfactual evidence differential introduced in the main paper. To make the analysis tractable, we consider the pairwise difference between the evidence differentials of the ground-truth class $y$ and a competing class $k$, which measures how the perturbation changes their relative prediction scores.

We proceed in three steps. First, we analyze the local behavior of this pairwise difference under small perturbations of the attention logits. Second, under a linear classifier assumption, we derive the sensitivity of the class logits with respect to the attention logits. Finally, we characterize the perturbations that maximize this quantity under the $L_1$ proximity constraint used in the proposed regularization.

\subsection{Local Analysis of the Evidence Margin}

Recall from Sec.~3.2 of the main paper that the counterfactual evidence differential is defined as
\begin{equation}
\Delta F(u,u^{\mathrm{cf}}) = F(u) - F(u^{\mathrm{cf}}),
\end{equation}
where $\Delta F_c(u,u^{\mathrm{cf}})$ denotes the component corresponding to class $c$.

\noindent
In the main paper, the loss $\mathcal{L}_{\mathrm{diff}}=\mathrm{CE}(\mathrm{softmax}(\Delta F),y)$ encourages the evidence differential of the ground-truth class $y$ to exceed those of competing classes. To analyze the local behavior of this objective, we consider the pairwise evidence margin between the ground-truth class $y$ and a competing class $k$:
\begin{equation}
m_{y,k}(u,u^{\mathrm{cf}})
: =
\Delta F_y(u,u^{\mathrm{cf}})
-
\Delta F_k(u,u^{\mathrm{cf}})
\end{equation}
Although $\mathcal{L}_{\mathrm{diff}}$ simultaneously enforces $\Delta F_y > \Delta F_k$ for all competing classes $k\neq y$, the cross-entropy objective can be written as $\mathcal{L}_{\mathrm{diff}} = \log\!\left(1+\sum_{k\neq y} e^{-m_{y,k}}\right)$, and  thus is primarily influenced by the most competitive class, i.e., the class with the largest $\Delta F_k$. Consequently, the optimization dynamics can be understood by analyzing the pairwise margin $m_{y,k}$ for a representative competing class. The following analysis therefore focuses on the
local behavior of $m_{y,k}$.

\noindent
Let
\begin{equation}
\delta = u^{\mathrm{cf}} - u
\end{equation}
denote a perturbation of the attention logits. In practice, $\delta$ quantifies the amount that must be injected into or removed from the instance attention logits in order to transform $u$ into $u^{\mathrm{cf}}$.
\noindent
Assuming that the class logits $F(u)$ are differentiable with respect to $u$, and let $c\in\{1,\dots,K\}$ denote a class index. A first-order Taylor expansion for sufficiently small $\delta$ yields:
\begin{equation}
\Delta F_c(u,u+\delta)
=
F_c(u)-F_c(u+\delta)
\approx
-\nabla_u F_c(u)^\top \delta
\end{equation}

\noindent
Therefore the local variation of the evidence margin satisfies:
\begin{equation}
m_{y,k}(u,u+\delta)
\approx
-
(\nabla_u F_y(u)-\nabla_u F_k(u))^\top \delta
\end{equation}

\noindent
Defining the margin sensitivity vector:
\begin{equation}
\label{eq:gyk}
g_{y,k}(u)
:=
\nabla_u F_y(u)-\nabla_u F_k(u)
\end{equation}
\noindent
we obtain therefore:
\begin{equation}
m_{y,k}(u,u+\delta)
\approx
-
g_{y,k}(u)^\top \delta
\end{equation}
This expression shows that the margin variation is governed by the sensitivity vector $g_{y,k}(u)$. In particular, perturbations aligned with $-g_{y,k}(u)$ increase the evidence margin, whereas perturbations
aligned with $g_{y,k}(u)$ decrease it.

\subsection{Attention Sensitivity under Linear Classifiers}

We now derive the explicit form of $\nabla_u F_c(u)$ when the classifier $\varphi$ is linear with respect to the aggregated bag representation.

Let $z_j\in\mathbb{R}^d$ denote the embedding of instance $j$, and let $w_c\in\mathbb{R}^d$ denote the classifier weights associated with class $c$. Assuming a linear classifier, the class-$c$ logit is:
\begin{equation}
F_c(u)=w_c^\top \hat{Z}
\end{equation}

\noindent
Using the bag representation defined in Eq.~(2) of the main paper,

\begin{equation}
\label{eq:lin_cl}
\hat{Z}=\sum_{j=1}^{N} a_j z_j \;\Rightarrow\;
F_c(u)=\sum_{j=1}^{N} a_j\, w_c^\top z_j
\end{equation}

\noindent
For convenience, we define $s_{c,j}$ in \cref{eq:scj} which can be interpreted as the class-$c$ logit that would be obtained if the bag contained only instance $j$. Substituting $s_{c,j}$ into \cref{eq:lin_cl}, the bag logit can be written as an attention-weighted average of instance-level logits:
\begin{equation}
\label{eq:scj}
s_{c,j}:=w_c^\top z_j 
\;\Rightarrow\;
F_c(u)=\sum_{j=1}^{N} a_j s_{c,j}
\end{equation}

\noindent
For convenience, we define in \cref{eq:diff_Fu} the quantity $\mu_c$, representing the attention-weighted average class alignment. Differentiating $F_c(u)$ with respect to the attention logits gives:
\begin{equation}
\label{eq:diff_Fu}
\mu_c := \sum_{j=1}^{N} a_j s_{c,j}, \qquad
\frac{\partial F_c}{\partial u_j}
= a_j (s_{c,j} - \mu_c).
\end{equation}

\noindent
Thus, the influence of instance $j$ on the class-$c$ logit depends on both its attention weight $a_j$ and the deviation of its instance-level logit (evidence) $s_{c,j}$ from the bag-level average $\mu_c$.

\subsection{Margin Sensitivity}

Using the class-wise attention sensitivity $\frac{\partial F_c}{\partial u_j}$ from \cref{eq:diff_Fu} and substituting it into \cref{eq:gyk}, the margin sensitivity along instance $j$ becomes:
\begin{equation}
g_{y,k,j}(u)
=
a_j\big[(s_{y,j}-s_{k,j})-(\mu_y-\mu_k)\big]
\end{equation}

\noindent
The term $(s_{y,j}-s_{k,j})$ measures how strongly instance $j$ supports class $y$ relative to class $k$, while $(\mu_y-\mu_k)$ represents the corresponding average evidence difference across the bag. Therefore, $(s_{y,j}-s_{k,j})-(\mu_y-\mu_k)$ measures how unusually discriminative instance $j$ is relative to the average evidence in the bag. Instances for which this quantity has large magnitude exert the strongest influence on the evidence margin.

\subsection{Optimal Perturbations under $L_1$ Distance}

We now analyze which perturbations of the attention logits maximize the evidence margin under an $L_1$ proximity constraint: $\|\delta\|_1\le \varepsilon$.
\noindent
Recall the first-order approximation above $m_{y,k}(u,u+\delta)\approx -g_{y,k}(u)^\top \delta$. So maximizing the margin reduces to:
\begin{equation}
\max_{\|\delta\|_1\le \varepsilon} -g_{y,k}(u)^\top \delta
\end{equation}

\noindent
This optimization follows from the dual relationship between the $L_1$ and $L_\infty$ norms:
\begin{equation}
\max_{\|\delta\|_1\le \varepsilon} g^\top \delta
=
\varepsilon \|g\|_\infty
\end{equation}
Hence the optimal perturbation concentrates on instances with maximal absolute margin sensitivity.
\noindent
Let:
\begin{equation}
M=\max_j |g_{y,k,j}(u)|,
\quad
S=\{j:\ |g_{y,k,j}(u)|=M\}
\end{equation}
Any optimal perturbation $\delta^\star$ must satisfy:
\begin{equation}
\mathrm{supp}(\delta^\star)\subseteq S,
\qquad
\|\delta^\star\|_1=\varepsilon,
\qquad
\mathrm{sign}(\delta^\star_j)=-\mathrm{sign}(g_{y,k,j}(u))
\quad \text{for } j\in S
\end{equation}
Thus the optimal perturbation primarily modifies the attention logits of the most discriminative instances. In particular, instances strongly supporting the ground-truth class yield positive margin sensitivity and therefore receive negative perturbations, reducing their attention in the counterfactual branch, whereas instances with negative margin sensitivity receive positive perturbations. \\

In the implementation used in the main paper, the $L_1$ proximity term is normalized by the bag size: $\mathcal{L}_{\mathrm{div}}= \frac{1}{N}\|u-u^{\mathrm{cf}}\|_1
$, which corresponds to the $L_1$ formulation in Eq.(9) of the main paper. This normalization keeps the scale of the regularization comparable across bags of different sizes, preventing large bags from inducing disproportionately large penalties and thereby improving training stability.\\

Overall, this analysis suggests that meaningful counterfactual explanations arise from small but structured redistributions of attention logits targeting instances with maximal margin sensitivity. This directly motivates the design of the training objective in the main paper: $\mathcal{L}_{\mathrm{diff}}$ encourages prediction differences between factual and counterfactual branches, while $\mathcal{L}_{\mathrm{div}}$ constrains their attention logits to remain close. Although the analysis above focuses on the $L_1$ case, the same principle also motivates the cosine-based proximity used in the main paper, which promotes directional differences between attention patterns while preserving their overall scale.

\section{Dataset Descriptions}
\label{supp_sect_5}
\subsection{WSI Datasets}
\cref{tab:supp_data} presents in more details the description of the different datasets used in our study. Details about the number of samples and classes are included, together with a small description of each of the digital pathology whole slide image data. More details can be identified in the original papers.
\begin{table}[ht]
\caption{\textbf{Overview of Whole-Slide Image (WSI) datasets used in this study.} For each cohort, we report the data source, clinical or diagnostic task, number of slides, and the distribution of labels.}
\label{tab:supp_data}
\centering
\resizebox{0.7\linewidth}{!}{
\begin{tabular}{p{2.5cm} p{4.5cm} c p{2.5cm}}
\toprule
\textbf{Cohort} & \textbf{Description} & \textbf{\# Slides} \hspace{0.2cm}  & \textbf{Labels with \#} \\

\midrule

\multicolumn{4}{c}{\textbf{TCGA} \cite{tomczak2015review}} \\
\midrule

BRCA & 
WSIs from TCGA-BRCA. Used for histological subtype classification between invasive ductal carcinoma (IDC) and invasive lobular carcinoma (ILC). 
& 977 & 
\begin{tabular}{@{}l@{}}IDC: 779 \\ ILC: 198\end{tabular} \\[2pt]

NSCLC & 
WSIs from TCGA-LUAD and TCGA-LUSC. Used for distinguishing between lung adenocarcinoma (LUAD) and lung squamous cell carcinoma (LUSC). 
& 956 & 
\begin{tabular}{@{}l@{}}LUAD: 478 \\ LUSC: 478\end{tabular} \\[2pt]

LUAD (TP53) & 
TCGA-LUAD dataset for TP53 mutation prediction directly from H\&E WSIs.  
Labels correspond to mutation status (mutated vs wild-type). 
& 427 & 
\begin{tabular}{@{}l@{}}TP53 WT: 199 \\ TP53 Mut: 228\end{tabular} \\

\midrule
\multicolumn{4}{c}{\textbf{BRACS (BACH Challenge)} \cite{BRACS}} \\ 
\midrule
BRACS \hspace{10mm} (7 classes) & 
Breast biopsy WSIs with detailed, fine-grained categories (7 diagnostic classes).  
& 546 & 
\begin{tabular}{@{}l@{}}Class 0: 44 \\ Class 1: 147 \\ Class 2: 73 \\ Class 3: 48 \\
Class 4: 41 \\ Class 5: 61 \\ Class 6: 132\end{tabular} \\

\midrule
\multicolumn{4}{c}{\textbf{Camelyon 16 Challenge} \cite{bejnordi2017diagnostic}} \\
\midrule

Camelyon16 (Train) & 
Lymph node metastasis detection dataset. Training subset of whole-slide images (H\&E) annotated for the presence of tumor metastasis. 
& 270 &
\begin{tabular}{@{}l@{}}Normal: 159 \\ Tumor: 111\end{tabular} \\[2pt]

Camelyon16 (Test) & 
Official test subset from the Camelyon16 challenge.
& 129 & 
\begin{tabular}{@{}l@{}}Normal: 80 \\ Tumor: 49\end{tabular} \\

\bottomrule
\end{tabular}
}
\end{table}

\subsection{Synthetic Datasets}

For the synthetic experiments, we reproduce the settings introduced in \cite{hense2025xmilinsightfulexplanationsmultiple}, including the \emph{Four Bags} and \emph{Adjacent Pairs} datasets.  
Each dataset is defined through an \emph{evidence function} $\varepsilon_j(c)$, which assigns to each instance $x_j$ and each class $c \in \{1,..,K\}$ where $K$ is the number of classes, a value in $\{-1, 0, 1\}$, indicating whether the instance provides \emph{negative} ($\varepsilon_j(c) = -1$), \emph{neutral}($\varepsilon_j(c) = 0$), or \emph{positive}($\varepsilon_j(c) = 1$) evidence for class $c$.

\myparagraph{Four Bags dataset}
Digits 8 and 9 provide opposite evidence for different classes.  
For an instance $x_j$, the evidence assignments are:

\begin{equation}
x_j \sim 8:\quad
\varepsilon_j(c) = 
\begin{cases}
1 & c \in \{1,3\},\\
-1 & c \in \{0,2\}\\
\end{cases}
\label{eq:evidence_8}
\end{equation}

\begin{equation}
x_j \sim 9:\quad
\varepsilon_j(c) = 
\begin{cases}
1 & c \in \{2,3\},\\
-1 & c \in \{0,1\}\\
\end{cases}
\label{eq:evidence_9}
\end{equation}

Otherwise :

\begin{equation}
\varepsilon_j(c) = 0  \quad \forall c \in \{0,1,2,3\}\\
% \label{eq:evidence_9}
\end{equation}

\myparagraph{Adjacent Pairs dataset}
Digits provide evidence only when specific pairs co-occur.  
Digit 4 supports class 1 and refutes class 0 \emph{only if digit 3 is present in the same bag}:

\begin{equation}
\label{eq:evidence_adj_pairs}
x_j \sim 4:\quad
\varepsilon_j(c) =
\begin{cases}
1  & \text{if } c=1 \text{ \& digit 3 is in the bag},\\[4pt]
-1 & \text{if } c=0 \text{ \& digit 3 is in the bag},\\[4pt]
0  & \text{otherwise}.
\end{cases}
\end{equation}

Evidence for other digits follows \cite{hense2025xmilinsightfulexplanationsmultiple}.

\section{Evaluation Details}
\label{supp_sect_6}

\subsection{Evaluation Metrics: AUPRC$^+$ and AUPRC$^\pm$}

Given a bag $B = \{\mathbf{x}_j\}_{j=1}^{N}$ and class $c \in \{1,..,K\}$, let the ground-truth evidence vector be:

\begin{equation}
e^{(c)} = \{\varepsilon_j(c)\}_{j=1}^{N}
\label{eq:evidence_vector}
\end{equation}

\noindent 
In most real-world histopathology datasets, such instance-level evidence is unavailable. However, for certain binary tasks such as metastasis detection on CAMELYON16\cite{bejnordi2017diagnostic} (Sec. 4 main paper), pixel-level tumor annotations allow patch-level labels to be derived. In this setting, the presence of tumor patches directly determines the bag label, making patch labels a well-defined proxy for positive evidence supporting the prediction.

\noindent We define the explanation score vector as:

\begin{equation}
s^{(c)} = \{s_j(c)\}_{j=1}^{N}
\label{eq:score_vector}
\end{equation}

\noindent where each score is computed as the sigmoid of the raw / non normalized attention value:

\begin{equation}
s_j(c) = \sigma\!\left(u_j\right)
\label{eq:score_sigmoid}
\end{equation}

\noindent To evaluate the explanation quality, we define binary targets for positive and negative evidence:

\begin{equation}
\label{eq:targets_pos_neg}
\begin{cases}
e^{(c)}_{\text{pos}} = \mathbf{1}\!\left[\varepsilon_j(c)=1\right] \\[4pt]
e^{(c)}_{\text{neg}} = \mathbf{1}\!\left[\varepsilon_j(c)=-1\right]
\end{cases}
\end{equation}

\noindent We compute the corresponding one-vs-all AUPRC scores:

\begin{equation}
\label{eq:auprc_pos_neg}
\left\{
\begin{aligned}
\mathrm{AUPRC}^{+} &= \mathrm{AUPRC}\!\left(e^{(c)}_{\text{pos}},\, s^{(c)}\right) \\[4pt]
\mathrm{AUPRC}^{-} &= \mathrm{AUPRC}\!\left(e^{(c)}_{\text{neg}},\, -s^{(c)}\right)
\end{aligned}
\right.
\end{equation}

\noindent We average these two values to obtain the AUPRC-2 metric:

\begin{equation}
\label{eq:auprc2}
\begin{aligned}
\mathrm{AUPRC^\pm}
= \frac{1}{2}\big(
&\mathrm{AUPRC}(e^{(c)}_{\text{pos}}, s^{(c)}) \\
&+\, \mathrm{AUPRC}(e^{(c)}_{\text{neg}}, -s^{(c)})
\big).
\end{aligned}
\end{equation}

\subsection{Perturbation Curves}

We qualitatively assess the quality of MIL models' attention as an interpretability proxy, following the region perturbation strategy used in~\cite{hense2025xmilinsightfulexplanationsmultiple,alber2019innvestigate}. For a bag $B$ containing $N$ patches, we sort instances by decreasing attention scores and partition them into $100$ disjoint groups
$r_1, \dots, r_{100}$, each containing $1\%$ of the most relevant instances according to attention values ( $r_1$ contains the top-$1\%$ most relevant patches, $r_2$ the next most relevant $1\%$, and so on).
\noindent
We construct perturbed versions of the bag by progressively removing the most relevant groups. Let $B^{(0)} = B$ denote the original slide. At step $k$, we drop the top $k\%$ most relevant patches and define the perturbed bag:

\begin{equation}
\label{eq:perturb_system}
\left\{
\begin{aligned}
B^{(k)} &= P(B, k) = \displaystyle\bigcup_{i = k+1}^{100} r_i,
\qquad k = 0, 1, \dots, 99 \\[8pt]
& B^{(100)} = \mathbf{0}
\end{aligned}
\right.
\end{equation}

\section{Implementation Details Across Different MIL Settings}
\label{supp_sect_7}
Our framework regularizes the \emph{pre-softmax} attention scores that determine how instance features are aggregated into a bag representation. For any MIL architecture exposing such scores, we introduce factual and counterfactual attention mechanisms producing logits $u$ and $u^{\mathrm{cf}}$. The two branches share the feature extractor and classifier, and the counterfactual objective is applied directly at the level of these raw attention scores.

Table~\ref{tab:complexity} reports the profiled computational complexity of baseline MIL architectures and their CAR-MIL variants. Complexity was measured with a dummy input of shape $1 \times 10000 \times 1024$, corresponding to a single bag containing $10000$ instances with $1024$-dimensional features. We report multiply-accumulate operations (MACs) in billions and the number of learnable parameters in millions. 
Overall, the additional cost remains small for attention-based MIL models and moderate for transformer-based models, which is expected given the higher computational cost of transformer-based architectures.

\begin{table}
\centering
\caption{Model complexity comparison between baseline MIL architectures and their CAR-MIL counterparts. The reported MACs quantify the cost of one forward pass whereas the parameter count reflects model size independently of the input.}

\label{tab:complexity}
\begin{tabular}{lcc}
\toprule
Model & MACs (G) & Params (M) \\
\midrule
ABMIL & 5.899 & 0.657 \\
CAR-ABMIL & 5.901 & 0.657 \\
\midrule
DSMIL & 5.318 & 0.594 \\
CAR-DSMIL & 5.908 & 0.659 \\
\midrule
TransMIL & 24.796 & 2.672 \\
CAR-TransMIL & 34.653 & 3.722 \\
\bottomrule
\end{tabular}
\end{table}

\myparagraph{ABMIL}
In the standard attention-based MIL setting (ABMIL)~\cite{ilse2018attentionbaseddeepmultipleinstance}, the model already produces a vector of pre-softmax instance scores. CAR-MIL therefore only introduces a parallel attention branch generating $u^{\mathrm{cf}}$, while the encoder and classifier remain shared. Since the aggregation mechanism is unchanged, the additional parameters correspond only to the extra attention projection, resulting in negligible computational overhead (~\cref{tab:complexity}).

\myparagraph{DSMIL}
DSMIL~\cite{li2021dualstreammultipleinstancelearning} computes an instance-to-class attention matrix prior to softmax normalization. Let
\begin{equation}
U_{j,c} = q(z_j)^\top q(z_{m_c})
\end{equation}
denote the attention score between instance $j$ and the class-specific critical instance $z_{m_c}$. CAR-MIL introduces a second query projection producing counterfactual scores
\begin{equation}
U^{\mathrm{cf}}_{j,c} = q_{\mathrm{cf}}(z_j)^\top q_{\mathrm{cf}}(z_{m_c}).
\end{equation}
Given the predicted bag class $\hat{y}$, we extract the corresponding column
\begin{equation}
u = U_{:,\hat{y}}, \qquad u^{\mathrm{cf}} = U^{\mathrm{cf}}_{:,\hat{y}}
\end{equation}
and apply the counterfactual objective to these vectors. Architecturally, this requires duplicating only the query projection used in the DSMIL bag classifier, leaving the remaining components unchanged. Consequently, the parameter increase is small and the computational overhead remains modest (Table~\ref{tab:complexity}).

\myparagraph{TransMIL}
Transformer-based MIL models such as TransMIL~\cite{shao2021transmiltransformerbasedcorrelated} compute attention through self-attention layers. In a standard transformer, attention derives from the pre-softmax query–key affinities
\begin{equation}
u = \frac{QK^\top}{\sqrt{d}},
\end{equation}
which would naturally align with our regularization. However, TransMIL employs Nystr\"om attention~\cite{xiong2021nystromformernystrombasedalgorithmapproximating}, which approximates the full attention matrix using landmark tokens and does not expose the exact pre-softmax attention logits governing bag aggregation.
\noindent
To integrate CAR-MIL, we therefore introduce a counterfactual transformer block parallel to the final attention layer and extract class-token attention maps as a surrogate attention signal. The factual and counterfactual predictions are then obtained from the two transformer branches. Because Nystr\"om attention approximates full attention, the resulting attention maps should be interpreted as approximate proxies rather than exact pre-softmax logits. This modification duplicates only the final transformer block, leading to a moderate increase in parameters and computation compared to ABMIL and DSMIL, as reflected in Table~\ref{tab:complexity}.

\end{document}